\documentclass{article}

 \usepackage[eandd, final]{neurips_2026}
\usepackage{float}

\usepackage[utf8]{inputenc} 
\usepackage[T1]{fontenc}    
\usepackage{hyperref}       
\usepackage{url}            
\usepackage{booktabs}       
\usepackage{amsfonts}       
\usepackage{nicefrac}       
\usepackage{microtype}      
\usepackage{xcolor}         
\usepackage{natbib}
\usepackage{graphicx}
\usepackage{amsthm}
\newtheorem{theorem}{Theorem}[section]
\newtheorem{definition}[theorem]{Definition}
\usepackage{amsmath}
\usepackage{enumitem}
\usepackage{subcaption}
\usepackage{wrapfig}

\title{Evaluating Compositional Generalization\\ in Transformers: The Role of \\Composition Equivalence and Module Coverage}

\author{%
  Purva Pruthi \\
University of Massachusetts Amherst\\
Amherst, MA, USA \\
  \texttt{ppruthi@cs.umass.edu} \\
  \And
  Andrew Yuan \\
University of Massachusetts Amherst\\
Amherst, MA, USA \\
  \texttt{awyuan@cs.umass.edu} \\
\And
  Alexander D’Amour \\
  Google DeepMind \\
Boston, Massachusetts, U.S.A \\
  \texttt{alexdamour@google.com} \\
\And
  David Jensen \\
University of Massachusetts Amherst\\
Amherst, MA, USA \\
  \texttt{jensen@cs.umass.edu} \\
}

\begin{document}

\maketitle

\begin{abstract}
Compositional generalization is an important form of out-of-distribution generalization, defined as the ability to train on some combinations of modules (e.g., functions such as \texttt{sort}, \texttt{filter}) and then generalize to unseen combinations. Despite substantial empirical work evaluating compositional generalization in transformers, findings have been inconsistent because evaluations were often conducted on a limited set of train-test splits, defined in an ad hoc manner. Through a systematic evaluation, we identify two key aspects of the train-test distribution that strongly affect performance: \textit{composition equivalence} and \textit{module coverage}. We show that the apparent performance of direct models (trained only on final outputs) results primarily from exploiting composition equivalences—different sequences of modules that reduce to identical end-to-end functions. When benchmarks eliminate these equivalences, performance drops significantly, showing their inability to generalize to compositions of known modules that produce novel end-to-end functions. We further demonstrate that performance degrades monotonically as module coverage shifts increase—when modules appear in different contexts between train and test—affecting both direct and step-by-step models (trained on intermediate and final outputs). These findings have concrete implications for benchmark design: benchmarks that include compositional equivalences can lead to significant overestimation of the generalization abilities of direct models, and benchmarks that include both module coverage violations and compositional equivalences can exacerbate shortcut learning. As a case study, we demonstrate the role of compositional equivalence on the generalization performance of the pre-trained model Llama3-8B on the code-output prediction task in the coding benchmark \href{https://livecodebench.github.io/}{LiveCodeBench}. 
\end{abstract}

\section{Introduction}
Compositional generalization—generalizing to unseen module combinations after training on a limited set—is a strong form of out-of-distribution generalization. This kind of compositional reasoning is relevant to many real-world tasks, such as predicting outputs of new software programs, planning novel action sequences, or constructing proofs from known operations. As transformer-based models are increasingly used in real-world applications, such as software development, robotics, and scientific discovery, it is useful to know the conditions under which they will reliably exhibit compositional generalization. 

\begin{figure}
    \centering
    \includegraphics[width=\textwidth]{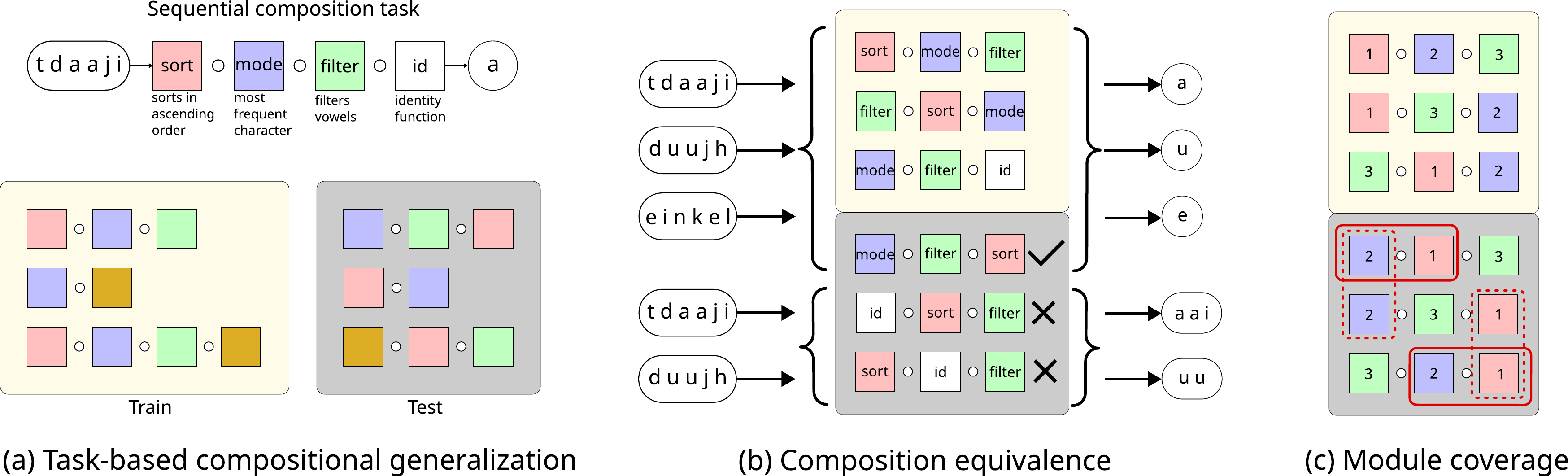}
    \caption{(a) A sequential composition task combines ``atomic'' functions (e.g., \texttt{sort}, \texttt{mode}, \texttt{filter}). \textbf{Task-based} compositional generalization is the ability to generalize to unseen sequences after training on a subset. (b) \textbf{Composition equivalence:} Different sequences are composition equivalent if they reduce to the same end-to-end function. For example, one equivalence class consisting of the first four sequences reduces to returning the most frequently occurring vowel in a set of strings; direct models succeed on the test sequence (\texttt{mode} $\circ$ \texttt{filter} $\circ$ \texttt{sort}) from this class but fail on the remaining two test sequences from a different composition equivalence class. (c) \textbf{Module coverage:} Examples of coverage failures include \textit{position-wise} (e.g., function 2 never appearing in the first position, shown with dashed lines) and \textit{pair-wise} (e.g., missing pair (2,1) in the training set, shown with solid lines).}
    \label{fig:intro_graphics}
\end{figure}

In this work, we focus on one of the simplest tasks in which compositional generalization can be evaluated: sequential compositional tasks. Such tasks consist of a sequence of ``atomic'' modules that transform input data into output data (see Figure \ref{fig:intro_graphics}(a) for an example), such as in software programs. We call this kind of generalization \textit{task-based compositional generalization}, since the input-output mapping task may differ as the sequence of modules changes.

Empirical evaluation of compositional generalization is more challenging than standard evaluation of in-distribution generalization because unseen compositions can lead to an unknown degree of shift between train and test distributions. In pre-trained large language models (LLMs), evaluation becomes even more challenging because the content of their training sets is largely unknown, making it difficult to determine whether high performance reflects compositional generalization or whether the test compositions were already seen in the training data. An additional challenge is that compositional settings admit a combinatorially large number of train-test partitions of the composition space, making exhaustive evaluation difficult~\citep{camposampiero2025scalable}. 

Due to these challenges, most existing empirical work ~\citep{hupkes2020compositionality, csordas2021devil, ontanon-etal-2022-making, dziri2023faith, song2025out, ramesh2024compositional} evaluates compositional generalization using a limited set of ad hoc train-test splits, often created via random partitions, which leads to inconsistent findings. For example, prior work~\citep{ramesh2024compositional, abedsoltan2025task, xu2024large} has shown that direct models (trained on final outputs) have limited generalization in certain settings, yet can perfectly generalize when the module function type or composition length is changed. Many existing explanations for this puzzling behavior focus on model-specific characteristics, such as insufficient computational depth~\citep{peng2024on}. 

In this work, we provide an alternative data-centric explanation, showing that significant variability in generalization performance can arise purely from the characteristics of benchmark design and how the train and test sets are created, irrespective of the model architecture. Through a systematic and controlled experimental design\footnote{Data and code to reproduce experiments are available at: \href{https://github.com/KDL-umass/compositional-generalization-transformers}{https://github.com/KDL-umass/compositional-generalization-transformers}} that allows precise control over the degree of distribution shift between train and test compositions, we identify two key factors that together explain the substantial variability observed across transformer variants—both direct (trained on final outputs) and step-by-step models (trained on intermediate outputs).

The first factor is \textit{\textbf{composition equivalence}} (Figure~\ref{fig:intro_graphics}(b))---where distinct module sequences reduce to the identical end-to-end function. Composition equivalence is a \textit{previously unknown} driver of compositional generalization. We observe that when train and test splits share such equivalences, direct models \textit{appear} to compositionally generalize, and performance is strongly predicted by the \textit{degree} of shared equivalences. We observe that the performance drops to \textit{near zero} when the overlap between composition equivalences in the training and test sets is eliminated. This implies that many positive results for direct models in the compositional generalization literature may be artifacts of benchmark design rather than evidence of compositional generalization. 

The second factor is \textit{\textbf{module coverage}} (Figure~\ref{fig:intro_graphics}(c))—the extent to which each atomic module appears at different positions and in similar contexts between train and test sets. We observe that as position-wise and pair-wise distributional gaps increase, generalization degrades for both direct models and step-by-step models. We observe that this effect is further amplified by composition equivalences, which encourage shortcut learning. Module coverage builds on the previously studied notion of data diversity \citep{ahuja2025on}, but we introduce a novel \textit{continuous} measure of coverage shift and study its interaction with composition equivalence. 

Having established a precise understanding of these factors in controlled settings with GPT-2-style models, we extend our analysis to larger pre-trained models (Gemma 3-1B and Llama 3-8B). We show that when composition equivalences are introduced into the fine-tuning set, compositional generalization performance on the code-output prediction task in \href{https://livecodebench.github.io/}{LiveCodeBench}—an existing coding benchmark—improves significantly, indicating that this effect extends to real-world LLM usage. These findings have concrete implications for both existing evaluations and future benchmark design. Most existing benchmarks define ``unseen'' compositions syntactically---whether a module sequence appeared in training---but we argue that a more appropriate definition is functional: if a test composition produces a novel end-to-end mapping as compared to the training compositions. 

\section{Related Work}
\label{sec:rel_works}
Our work addresses key gaps in understanding the conditions under which compositional generalization occurs in transformers. An additional discussion of related work is provided in~\ref{ref:appendix:extended_rel_works}.

\textbf{Systematic evaluation of compositional generalization in transformers:} The closest framework to ours is \citet{ramesh2024compositional}. Our work differs in two key ways. First, they focus primarily on a single module type (a random bijection, similar to a one-to-one lookup table), whereas we include modules with varied functional properties more representative of real-world compositional reasoning tasks (e.g., program execution in software), through which we discover the novel factor of composition equivalence. Second, they combine within-$k$ and cross-$k$ evaluations (whether the composition length $k$ differs between train and test) using identity modules as padding, whereas we separate these settings and show that performance differs substantially between them. Our findings on composition equivalences explain the limited direct model generalization observed in their work, as well as why direct model performance improves across different module function types~\citep{xu2024large}. \citet{chang2025coverage} also discusses functional equivalence and coverage principles, seemingly related concepts to composition equivalence and module coverage, but focuses on \textit{data-based} generalization---novel data combinations within a \textit{fixed} task---whereas we study \textit{task-based} generalization. Consequently, their functional equivalence operates at the input \textit{data} level, while composition equivalence operates at a higher abstraction, relating compositions by their shared end-to-end \textit{function}.

\textbf{Relation of module coverage to prior work:} Module coverage is related to previously studied concepts of compositional support \citep{wiedemer2023compositional}, connected support \citep{schug2024discovering, redhardt2025scaling}, and sufficient data diversity \citep{ahuja2025on}, required for successful compositional generalization across different domains (e.g., vision). These requirements state that the training distribution should include non-zero position-wise support for each module or module combination and, in general, sufficiently diverse combinations. Prior work typically assumes uniform sampling across train and test to ensure diversity, thereby covering only a limited range of distribution shifts. Task-based compositional generalization work \citep{ramesh2024compositional, abedsoltan2025task} studied model sensitivity to extreme position-wise support violations. Our formulation provides a \textit{continuous} metric that captures a broader range of coverage shifts, spanning from uniform sampling to extreme support violations. We also provide a systematic procedure for constructing train/test splits with varying degrees of module-coverage-based shifts (Section~\ref{subsec:creating_splits}). 

\section{Distribution shifts characterizing task-based compositional generalization} 
\label{sec:distribution_shifts}
Consider a set of $n$ module functions $\mathcal{F} = \{f_1, \ldots, f_n\}$, where each $f_i: \mathcal{V}^m \rightarrow \mathcal{V}^m$ maps input sequences to output sequences over vocabulary $\mathcal{V}$. A sequential composition of length $k$ is an ordered sequence $(f_1, \ldots, f_k)$ applied as $\mathbf{y} = (f_k \circ \cdots \circ f_1)(\mathbf{x})$, with intermediate outputs $\mathbf{y}_j = (f_j \circ \cdots \circ f_1)(\mathbf{x})$. Compositions are represented using module tokens $\mathcal{T} = \{t_1, \ldots, t_n\}$, with input sequence $(t_1, \ldots, t_k, x_1, \ldots, x_m)$; models are trained auto-regressively to predict the output. Data tokens are sampled uniformly from $\mathcal{V}^m$ in both train and test distributions.

\begin{definition}[\textbf{Task-based compositional generalization}]
Let $\mathcal{T}^k$ denote the space of module sequences of length $k$. Task-based compositional generalization is evaluated via train-test distributions with disjoint supports: $\text{supp}(\mathcal{P}_{\text{train}}^{\mathcal{T}}) \cap \text{supp}(\mathcal{P}_{\text{test}}^{\mathcal{T}}) = \emptyset$, where training sequences are drawn from $\Omega \subset \mathcal{T}^k$ and test sequences from $\mathcal{T}^k \setminus \Omega$.
\end{definition}

\subsection{\textbf{Composition equivalence based distribution shift}}
\label{subsec:comp_eq}

Two distinct sequential compositions are \textit{equivalent} if they induce the same input-output mapping over a given set of data tokens.

\begin{definition}[\textbf{Composition equivalence and equivalence class}]
Two compositions $F = f_1 \circ \cdots \circ f_k$ and $F' = f'_1 \circ \cdots \circ f'_k$ are \textbf{equivalent} over input subspace $\mathcal{S} \subseteq \mathcal{V}^m$ if $F(\mathbf{x}) = F'(\mathbf{x})$ for all $\mathbf{x} \in \mathcal{S}$. The \textbf{equivalence class} of $F$ is:
$$[F]_{\mathcal{S}} = \{F' : F'(\mathbf{x}) = F(\mathbf{x}) \;\forall\, \mathbf{x} \in \mathcal{S}\}$$
\end{definition}

Some simple examples of composition equivalence include:
\begin{enumerate}
    \item \textbf{Identity-based exact equivalence:} For any composition $F = f_1 \circ \cdots \circ f_k$, inserting $\text{id}$ at any position yields an equivalent sequence: $F_0 := f_1 \circ \cdots \circ f_k \circ \text{id}$, \ldots, $F_k := \text{id} \circ f_1 \circ \cdots \circ f_k$.

    \item \textbf{Approximate equivalence due to module properties:} Consider \texttt{filter} (extracts vowels), \texttt{mode} (most frequent character), and \texttt{sort} (lexicographic order). For strings where the most frequent character is a vowel, \texttt{sort(mode(filter($\cdot$)))} and \texttt{filter(sort(mode($\cdot$)))} are equivalent (Figure~\ref{fig:intro_graphics}(b)).

    \item \textbf{Different implementations of the same function:} Recursive and iterative implementations of the Fibonacci sequence.
\end{enumerate}

\begin{definition}[\textbf{Degree of shared composition equivalences}]
The degree of shared equivalences between train and test is the proportion of test compositions that share an equivalence class with at least one training composition:
$$\delta_{\text{equiv}} = \frac{|\{F_B \in B : \exists\, F_A \in A \text{ s.t. } [F_B]_{\mathcal{S}}^g = [F_A]_{\mathcal{S}}^g\}|}{|B|}$$
\end{definition}

Conversely, $1 - \delta_{\text{equiv}}$ quantifies the extent to which test compositions exhibit novel end-to-end behavior unseen during training. While task-based compositional generalization is defined syntactically in terms of unseen module combinations, composition equivalence captures the semantic novelty of unseen compositions. 

\subsection{Module coverage based distribution shift}
\label{subsec:mod_coverage}
Module-coverage-based distribution shift quantifies how frequently atomic modules appear in specific positions or contexts at test time relative to training. Such shifts can introduce spurious correlations between positions and modules, causing generalization failures when these correlations break. We define two metrics capturing KL divergence in the \textit{position-wise} and \textit{pair-wise} module distributions, as KL divergence is asymmetric and measures shifts in the test distribution relative to training. 

\begin{definition}[\textbf{Position-wise and pairwise module coverage shift}]
The empirical position-wise and pairwise distributions are: $$\hat{P}(T_j = t_i) = \frac{1}{N}\sum_{s=1}^N \mathbb{I}(T_j^{(s)} = t_i), \quad \hat{P}(T{=}t_i, T'{=}t_{i'}) = \frac{1}{N(k-1)}\sum_{s=1}^N\sum_{j=1}^{k-1} \mathbb{I}(T_j^{(s)}{=}t_i,\, T_{j+1}^{(s)}{=}t_{i'})$$ $$D_{\text{pos}} = \frac{1}{k}\sum_{i=1}^k D_{KL}(\hat{P}_{\text{test}}(T_i) \,\|\, 
\hat{P}_{\text{train}}(T_i)), \quad D_{\text{pair}} = D_{KL}(\hat{P}_{\text{test}} (T,T') \,\|\, \hat{P}_{\text{train}}(T,T'))$$
\end{definition}

\subsection{Creating benchmarks with a controllable degree of distribution shifts}
\label{subsec:creating_splits}
We first describe two data sets that differ in their composition equivalences, and then describe how we create benchmarks with varying degrees of shift.
\subsubsection{Data sets}
The \textit{uniform} data set eliminates composition equivalence by default, serving as a control, while the \textit{diverse} data set captures the richer equivalence structure of real-world functions.

\textbf{Uniform data set:} We use six random bijection functions, each mapping input tokens to output tokens via a one-to-one lookup table~\citep{ramesh2024compositional}. When mappings are unique, non-identity, and uniformly random, any $k$-module composition produces a \textbf{unique} labeling function with high probability. This implies that no two composition sequences are equivalent. However, similar labeling functions are still possible in rare cases; for example, two bijections can be inverses of each other, leading to an identity labeling function. We empirically validate uniqueness by comparing outputs of every pair of compositions on a large number of random strings (1000 strings), and we find no duplicate labeling functions. Adding an identity module introduces exact identity-based equivalence, allowing us to evaluate the effect of a complete lack of equivalence versus exact equivalence in a controlled setting. 

\textbf{Diverse data set:} We create a dataset of string-manipulation functions inspired by RASP primitives~\citep{weiss2021thinking, zhou2024what} that transformers can represent and learn. We use simple functions to ensure observed limitations reflect compositional reasoning failures rather than insufficient model capacity. The six module functions---\texttt{sort, concatenate, filter, shift, union, mode}---are defined as follows: \texttt{shift} maps each character one position forward (a$\rightarrow$b, z$\rightarrow$a); \texttt{sort} rearranges characters lexicographically; \texttt{mode} returns the most frequent character (ties broken lexicographically); \texttt{concatenate} appends the second string to the first; \texttt{union} returns unique characters in order of first occurrence; and \texttt{filter} extracts vowels in order. We pad outputs with \texttt{<SPACE>} to maintain fixed length $m$. These functions induce composition equivalences: \texttt{filter} and \texttt{mode} create many-to-one mappings enabling shortcut reasoning; \texttt{concatenate} and \texttt{union} produce identical outputs for strings with distinct characters; and \texttt{shift} and \texttt{sort} are commutative for many strings. These properties result in both exact and approximate equivalences (Figure~\ref{fig:equivalence_class_visualization}). 

\subsubsection{Constructing train-test splits with varying degrees of composition equivalences and module coverage}
\label{sec:creating_splits}
\textbf{Composition equivalences:} In the diverse benchmark, equivalence classes are not pre-defined as in the case of the uniform benchmark. Therefore, we learn equivalence classes from data by using the following approach.

\begin{itemize}
    \item Compute the pairwise composition equivalence score $CE(F, F') = \frac{1}{N}\sum_{s=1}^{N} \mathbb{I}[F(\mathbf{x}) 
= F'(\mathbf{x})]$ between two task sequences as the proportion of inputs for which the final outputs match when evaluated over a large number of $N=1000$ inputs sampled from the vocabulary.
    \item Construct an undirected graph based on the equivalence scores, where each node denotes a composition sequence and an edge denotes the score strength above a threshold of $0.01$. That is, two compositions are considered equivalent if they produce the same outputs in at least 1\% of cases. We chose this low, conservative threshold so that even weak composition equivalences are captured when learning equivalence classes from data. 
    \item Learn equivalence classes by finding connected components in the undirected graph. 
    \item \textbf{Disjoint splits ($\delta_{\text{equiv}} = 0$):} Split train and test data by equivalence class---no class appears in both sets. We consider 50-50 splits to accommodate varying degrees of equivalence, as explained below.
    \item \textbf{Splits with varying degrees of composition equivalences ($\delta_{\text{equiv}} \in (0,1]$)}:  Increase shared equivalences by swapping roughly half of the members of each test equivalence class with members from a training class. This systematically increases test sequences with shared equivalences in proportion to class sizes, while maintaining a constant test set size.
\end{itemize}
\textbf{Module coverage:} We start with uniform position-wise and pair-wise distributions $P_{\text{train}}(T_j = t_i) = P_{\text{test}}(T_j = t_i) = \frac{1}{n}$ and introduce a bias strength parameter $\alpha \in [0,1]$ that skews sampling probabilities in complementary directions for train and test at each position. At $\alpha=0$, distributions remain uniform; as $\alpha \to 1$, the position-wise KL divergence $D_{\text{pos}}$ increases monotonically, creating larger coverage shifts while maintaining disjoint train-test sets. See Section~\ref{app:mod_coverage_shift} for details.

\subsection{Evaluation setup}
\label{subsec:evaluation_setup}
\textbf{Within-$k$ and cross-$k$ generalization:} In \textit{within-$k$}, train and test are created using randomly sampled disjoint  80/20 splits of $\mathcal{T}^k$ for $k \in \{2,\ldots,6\}$, where $k$ is the number of non-identity modules. In \textit{cross-$k$}, models trained on $\mathcal{T}^k$ are evaluated on $\mathcal{T}^{k'}$ for $k \neq k'$, with identity modules used as padding to align sequence lengths.

To enable a fair comparison between the two settings and assess model sensitivity to identity modules, we evaluate within-$k$ in two variants: (1) \textit{without identity}, where no identity module is used; (2) 
\textit{with identity}, where compositions are padded with identity modules to a fixed length of 7, mirroring cross-$k$. For example, in the without-identity variant, a within-$2$ model trained 
on $\{\texttt{shift} \circ \texttt{sort}\}$ is evaluated on $\{\texttt{sort} 
\circ \texttt{shift}\}$. In the with-identity variant, the same model is 
trained on $\{\texttt{id} \circ \texttt{shift} \circ \texttt{sort} \circ 
\texttt{id} \circ \texttt{id} \circ \texttt{id} \circ \texttt{id}\}$ and 
evaluated on $\{\texttt{id} \circ \texttt{sort} \circ \texttt{shift} \circ 
\texttt{id} \circ \texttt{id} \circ \texttt{id} \circ \texttt{id}\}$. A cross-$k$ model trained on $k=6$ compositions such as $\{\texttt{filter} 
\circ \texttt{sort} \circ \texttt{mode} \circ \texttt{shift} \circ 
\texttt{union} \circ \texttt{id} \circ \texttt{concatenate}\}$ is evaluated 
on $k'=3$ compositions such as $\{\texttt{id} \circ \texttt{filter} \circ 
\texttt{sort} \circ \texttt{id} \circ \texttt{mode} \circ \texttt{id} \circ 
\texttt{id}\}$.

\textbf{Models:} We train four transformer variants---direct and step-by-step models, each with absolute and relative positional embeddings~\citep{shaw2018self}---using the nanoGPT architecture~\citep{karpathy2023nanogpt} (3 layers, 6 heads). We use this small architecture for efficient evaluation across many train-test splits; larger models (12 layers, 12 heads) show similar generalization trends (Appendix~\ref{app:larger_gpt2}). Training uses $100$k samples, and testing uses $10$k samples uniformly sampled per composition, with input tokens of length six sampled from $\mathcal{V} = \{a, \ldots, z\}$. Models are trained for 100--200 epochs until training loss saturates, achieving 100\% training accuracy and 90--100\% validation accuracy on unseen input strings within training compositions (see Appendix~\ref{app:experimental_details}).

\textbf{Evaluation criteria:} We report the mean accuracy on unseen sequences using exact-match scoring (1.0 for perfect predictions and 0.0 otherwise), averaged over five random seeds.

\section{Explaining compositional generalization performance via composition equivalence and module coverage}
\label{sec:findings}
We first discuss results that highlight the interplay between composition equivalence and module coverage in \textit{within-$k$} and \textit{cross-$k$} settings. We then isolate the individual role of each factor using controlled splits created in Section~\ref{sec:creating_splits}, and show that the degree of each shift strongly predicts the performance of both direct and step-by-step models.

\subsection{Direct models show near-perfect generalization in some settings but fail completely in others due to composition equivalences}
\label{sec:results_equivalence}

\begin{figure}[!ht]
    \centering
    \includegraphics[width=\textwidth]{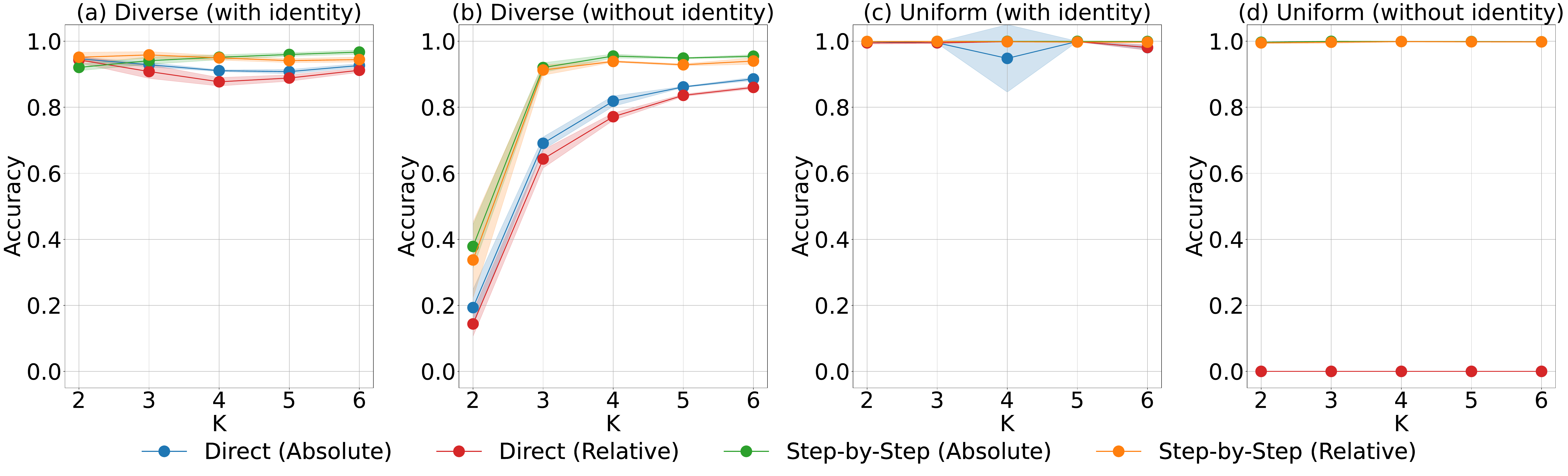}
    \caption{\textbf{Within-$k$ evaluation:} Direct models fail completely (0\% accuracy) on uniform bijections without identity tokens (Figure (d)), while step-by-step models maintain near-perfect performance across all conditions. We observe that including identity tokens or diverse function types seems crucial for the generalization performance of direct models.}
    \label{fig:identity_experiment}
\end{figure}

\begin{figure}[!ht]
    \centering
    \includegraphics[width=\textwidth]{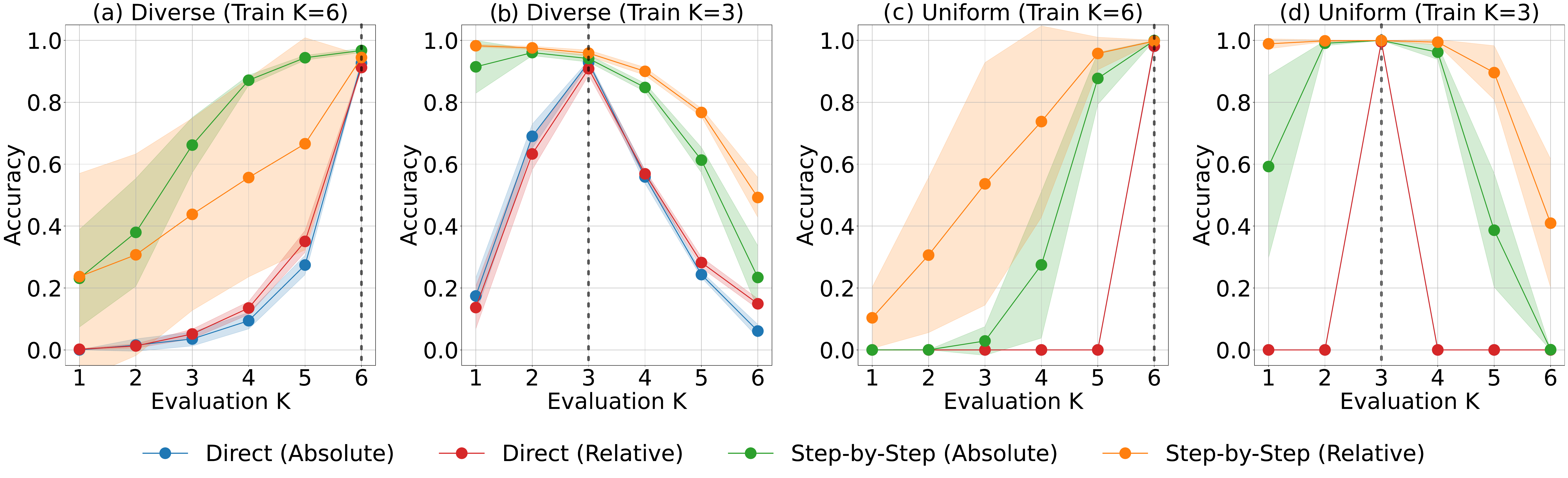}
    \caption{\textbf{Cross-$k$ evaluation:} For the diverse benchmark (a--b), direct models perform well near training $k$ as equivalences persist across nearby $k$ values. Direct models fail on the uniform benchmark (c--d) regardless of identity tokens, since identity-based equivalences in the uniform benchmark are $k$-specific. Step-by-step models are more robust but still exhibit significant variability.}
    \label{fig:cross_k_evaluation}
\end{figure}

In Figures~\ref{fig:identity_experiment} and \ref{fig:cross_k_evaluation}, direct models show a surprising range of performance variability: near-perfect generalization (70--100\% accuracy) in some settings, and complete failure (near-zero) in others, even for the simplest $k=2$ compositions (see Figure \ref{fig:identity_experiment}(d)). As we show below, this variability can be explained by the presence or absence of shared composition equivalences in the train-test split.

\textbf{Cases where direct models show significant generalization:}

\begin{itemize}
    \item \textbf{Diverse benchmark with or without identity modules (within-$k$):} Random 80/20 splits with identity modules contain \textit{exact} identity-based equivalences, resulting in near-perfect performance (Figure~\ref{fig:identity_experiment}a). Without identity modules (Figure~\ref{fig:identity_experiment}b), function properties produce \textit{exact} and \textit{approximate} equivalences (e.g., \texttt{sort} $\circ$ \texttt{mode} $\equiv$ \texttt{mode} $\circ$ \texttt{sort}) that direct models exploit, resulting in significant performance --- especially for $k=4$--$6$.

    \item \textbf{Uniform benchmark with identity modules (within-$k$):} Identity modules introduce exact equivalence classes, again resulting in near-perfect performance (Figure~\ref{fig:identity_experiment}c).

    \item \textbf{Diverse benchmark near training $k$ (cross-$k$):}
    Function-property-based equivalences exist across nearby $k$ values (e.g., (\texttt{sort} $\circ$ \texttt{filter}), (\texttt{sort} $\circ$ \texttt{filter} $\circ$\texttt{mode}) are equivalent for many input strings), explaining non-zero performance when evaluation $k$ is close to training $k$ (Figures~\ref{fig:cross_k_evaluation}a--b).
\end{itemize}

\textbf{Cases where direct models fail completely:}
\begin{itemize}
    \item \textbf{Uniform benchmark without identity modules (within-$k$):} No composition equivalences exist by construction, causing performance to drop to \textit{exactly zero} across all $k \in \{2, \ldots, 6\}$.

    \item \textbf{Uniform benchmark (cross-$k$):} Identity-based equivalences in uniform benchmark are $k$-specific---changing the number of non-identity modules changes end-to-end behavior---so they do not carry over across $k$, causing complete failure (Figures~\ref{fig:cross_k_evaluation}c--d).

    \item \textbf{Diverse benchmark far from training $k$ (cross-$k$):} As evaluation $k$ moves further from training $k$, function-property-based equivalences reduce due to changing end-to-end functions, and module coverage shifts also increase due to the varying distribution of identity modules, causing performance to drop sharply towards zero e.g., $k$=$1$--$4$ (Figure~\ref{fig:cross_k_evaluation}a), and $k$=$1$, $5$, and $6$. (Figure~\ref{fig:cross_k_evaluation}b)
\end{itemize}

In summary, direct models fail to generalize to compositions that produce novel end-to-end functions. 

\begin{figure}[!ht]
    \includegraphics[width=\textwidth]{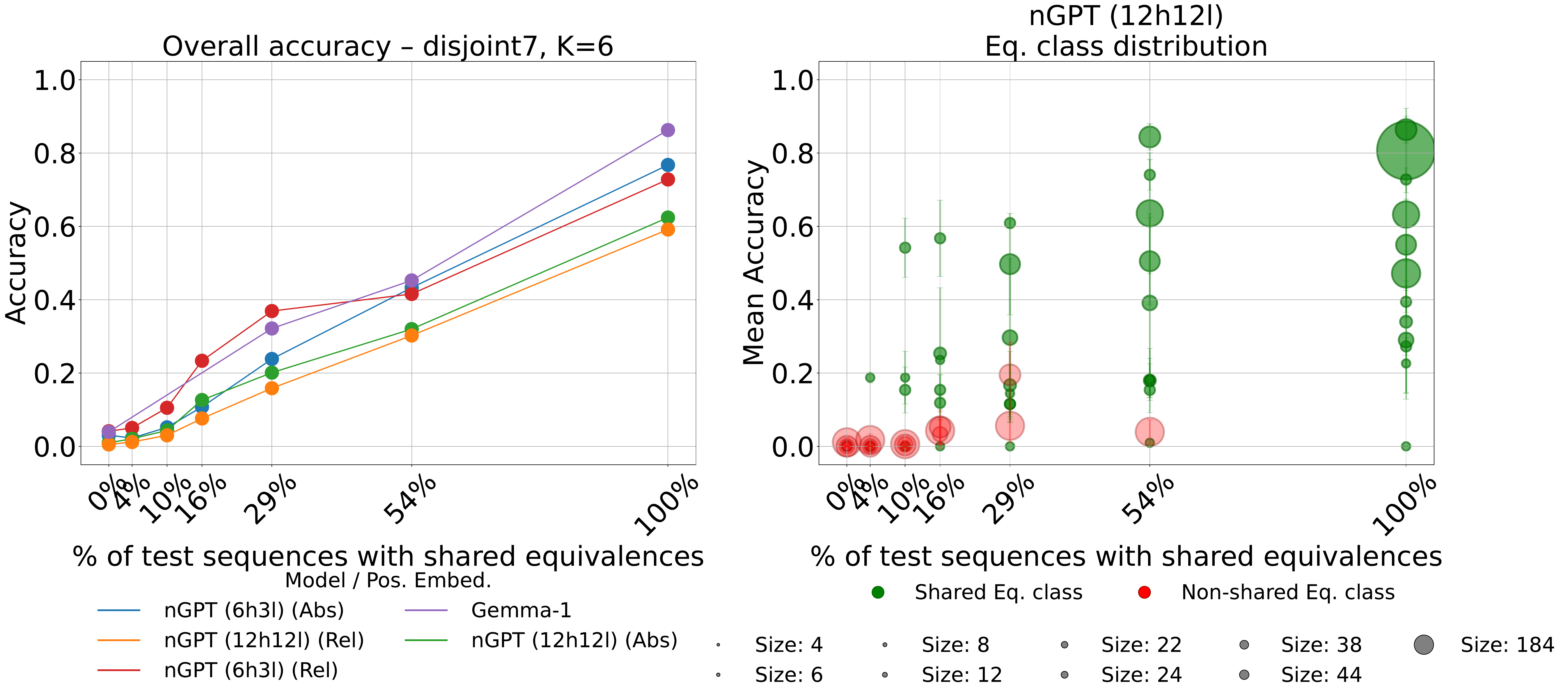}
    \caption{\textbf{Isolating the effect of composition equivalence in the diverse benchmark ($k=6$):} (a) Generalization performance increases with the degree of shared composition equivalence classes between train and test. (b) The direct model generalizes primarily to test sequences that share equivalence classes and fails on classes exhibiting novel end-to-end behavior.}
    \label{fig:equivalence_class_leakage_experiment_diverse}
\end{figure}

\begin{table}[t]
\centering
\caption{\textbf{Sensitivity of direct-model accuracy to the equivalence threshold:} We report accuracy on train/test splits with 0\% and 100\% shared composition equivalences, where equivalence classes are learned at different thresholds. At low thresholds, weak equivalences are captured, so the 0\% shared split contains no equivalences of any strength and generalization is poor. At high thresholds, some equivalences go undetected and remain in the 0\% shared split, which inflates generalization performance.}
\label{tab:threshold_sensitivity}
\small
\setlength{\tabcolsep}{6pt}
\begin{tabular}{lcc}
\toprule
\textbf{Threshold} & \textbf{Accuracy: 0\% shared} & \textbf{Accuracy: 100\% shared} \\
\midrule
0.01 & 0.04 & 0.70 \\
0.25 & 0.08 & 0.73 \\
0.50 & 0.28 & 0.84 \\
0.75 & 0.68 & 0.88 \\
0.95 & 0.84 & 0.91 \\
\bottomrule
\end{tabular}
\end{table}
\paragraph{Composition equivalence is the primary mechanism through which direct models
achieve compositional generalization:}

Figure~\ref{fig:equivalence_class_leakage_experiment_diverse}(a) shows that accuracy is strongly predicted by the degree of shared equivalences, rising sharply from 0\% (no sharing) to roughly 80\% (full sharing)---an effect that also holds for fine-tuned Gemma-3-1B models, confirming it is not limited to small GPT-2-style transformers. We observe that performance does not reach 100\% even with complete sharing, since some equivalences are approximate for a subset of inputs and some classes have only 4--6 members, making training subsets insufficient for perfect generalization. Figure~\ref{fig:equivalence_class_leakage_experiment_diverse}(b) further shows that accuracy is substantially higher for shared classes than non-shared ones, and that even some shared classes have lower performance, especially when class size is small. Together, these results establish that shared equivalence classes are \textit{necessary} but not \textit{sufficient} for compositional generalization in direct models. We observed similar results for the uniform benchmark (see Figure~\ref{fig:equivalence_class_leakage_experiment} in the Appendix). 
Our findings on composition equivalence also explain findings in~\citet{ramesh2024compositional}: their limited but not \textit{near-zero} generalization performance for direct models on the uniform benchmark (Figure 6 in ~\citet{ramesh2024compositional}) is likely a result of reporting accuracy over both train and test compositions rather than test alone and combining within-$k$ and cross-$k$ evaluations with identity padding, which inadvertently introduces equivalences.

\textbf{Sensitivity of the effect of shared composition equivalences to the threshold value:} We run the composition equivalence leakage experiment using equivalence classes learned with a threshold equivalence score of 0.01. To assess how sensitive the results are to higher thresholds, we also analyze accuracy at thresholds of 0.25, 0.5, 0.75, and 0.95. Results are presented in Table~\ref{tab:threshold_sensitivity}. At every threshold, we observe that accuracy increases monotonically from 0\% to 100\% shared equivalence. However, higher thresholds \textit{underestimate} true equivalence sharing, which \textit{overestimates} apparent generalization for a given shared equivalence value. For example, at threshold=0.95, the nominally 0\% shared condition has an accuracy of 0.85, whereas at threshold=0.01 the same condition results in an accuracy of 0.01. Compositions that share only weak equivalences (those producing matching outputs on fewer than 95\% of inputs) are placed in separate classes under threshold=0.95. As a result, the 0\% shared’’ condition at this threshold still contains test compositions that share undetected functional equivalences with the training data, and part of the 0.85 accuracy stems from these undetected equivalences. The low threshold of 0.01 avoids this inflation: its ``0\% shared’’ condition consists of disjoint classes with no shared equivalence, whether weak or strong.

\subsection{Module coverage shifts affect both direct and step-by-step models}
\label{sec:results_coverage}

Step-by-step models exhibit robust generalization across all splits in within-$k$ evaluation (Figure~\ref{fig:identity_experiment}) but perform well only near the training $k$ in cross-$k$ evaluation (Figure~\ref{fig:cross_k_evaluation}), showing significant variability. We observe that this is primarily due to the position-wise and pair-wise module distributions shifting substantially as the evaluation $k$ moves away from training $k$ (see divergence values in Figure~\ref{fig:cross_k_divergence} in the Appendix). 

\begin{wrapfigure}{r}{0.5\textwidth}
    \centering
    \includegraphics[width=0.48\textwidth]{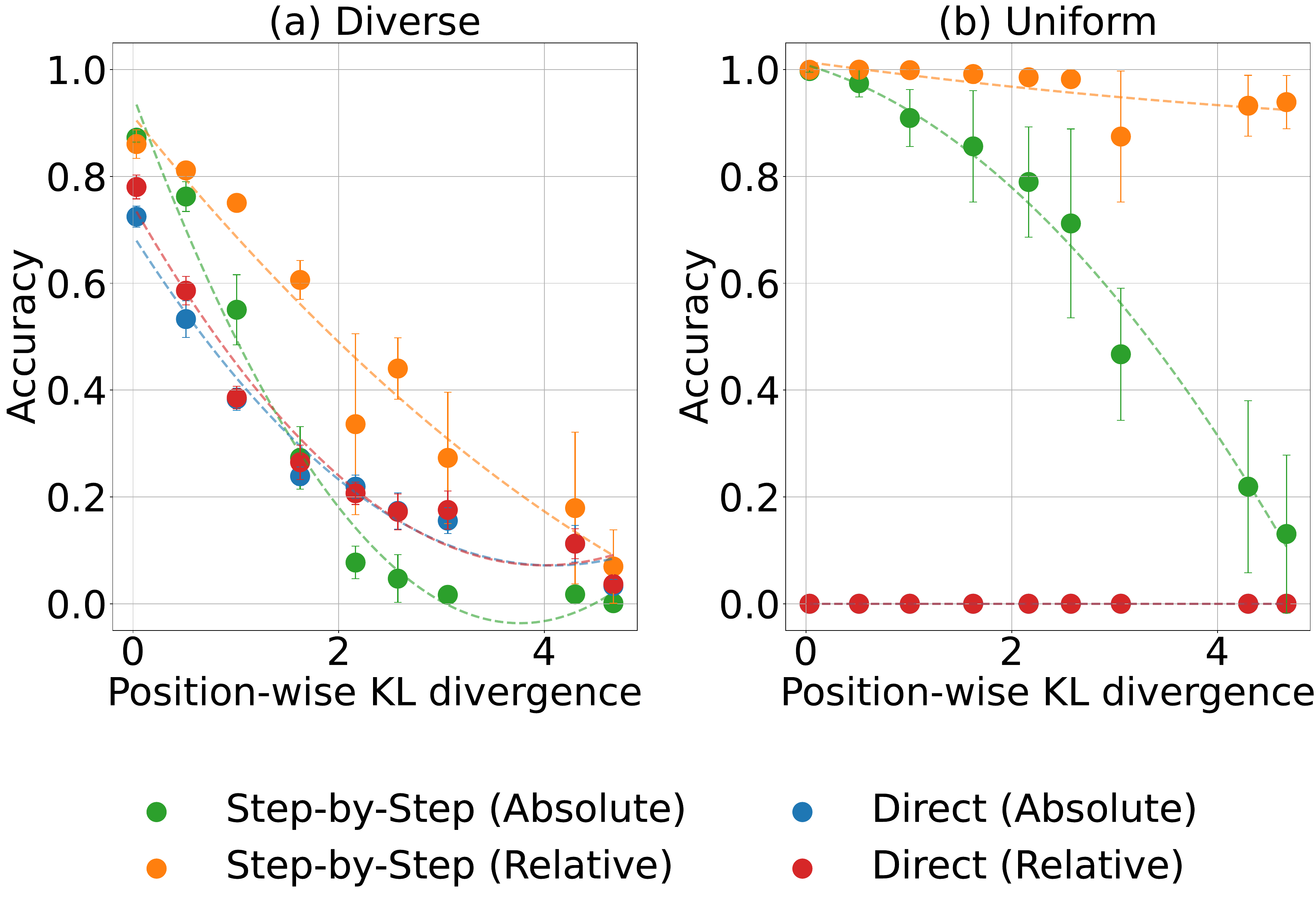}
    \caption{\textbf{Effect of position-wise distribution shift:} As the position-wise KL divergence increases, the compositional generalization performance of both direct and step-by-step models decreases.}
    \label{fig:position_wise_kl}
\end{wrapfigure}

\paragraph{Effect of module coverage shifts and interaction between module coverage and composition equivalence:}
Compositional generalization performance degrades monotonically as position-wise KL divergence increases across both benchmarks and model variants, indicating that models primarily rely on uniform module sampling for successful generalization (Figure~\ref{fig:position_wise_kl}). On the diverse benchmark, the presence of composition equivalences accelerates this degradation for all models (Figure~\ref{fig:position_wise_kl}(a)), as equivalences encourage shortcut learning that compounds coverage-based failures. In step-by-step models, absolute position encodings are more sensitive to position-wise shifts than relative encodings, particularly on the uniform benchmark (Figure~\ref{fig:position_wise_kl}(b)). For pairwise distribution shifts (Figure~\ref{fig:pairwise_kl} in the Appendix), models are more sensitive to the diverse benchmark than the uniform benchmark, again due to equivalence-driven shortcuts; on the uniform benchmark, performance remains robust except in the case of extreme support violations.

Finally, Figure~\ref{fig:equivalence_class_leakage_experiment}(a) (Appendix) reveals an additional interaction between the two factors: direct models with absolute position embeddings fail to learn correct equivalences, mainly due to module coverage failures (e.g., test compositions always starting with \texttt{id}). This shows that composition equivalence and module coverage are \textit{not independent}---a dependence that naturally arises in real-world settings where certain module orderings are more prevalent, producing more or fewer equivalences.

\subsection{Case Study: Composition equivalence in Llama3-8B model on \href{https://livecodebench.github.io/}{LiveCodeBench}}
\begin{figure}
    \centering
    \includegraphics[width=\linewidth]{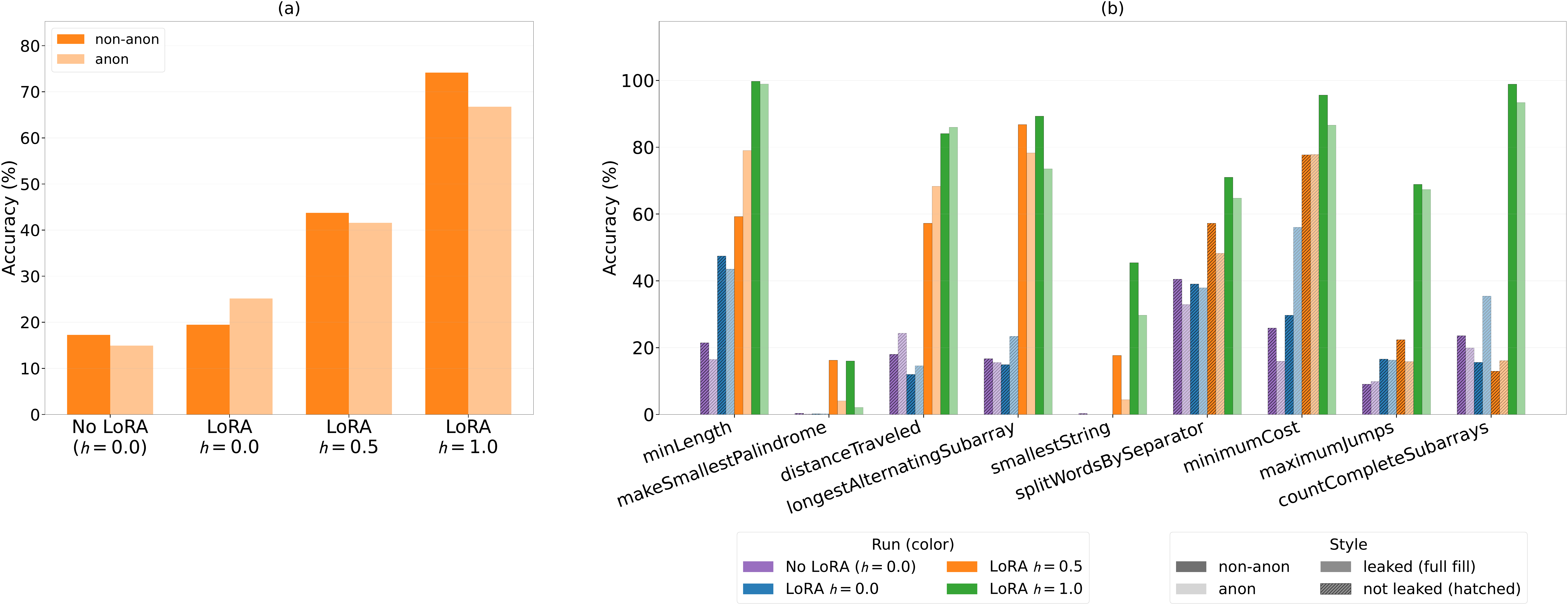}
    \caption{(a) \textbf{Code output prediction performance of Llama3-8b on LiveCodeBench benchmark:} We show the accuracy of the base model (No LoRA) and three LoRA fine-tuned models on splits with varying ratios of shared equivalence classes (defined by the same function names) (held out $h$=0, 0.5, 1.0). As splits share composition equivalences, accuracy improves substantially. (b) \textbf{Performance grouped by equivalence classes:}  We observe that performance on \textit{most} functions only improves when equivalences are leaked into the fine-tuning set.}
    \label{fig:LiveCodeBench_combined}
\end{figure}

In Figure~\ref{fig:LiveCodeBench_combined}, we present the performance of the Llama3-8B model on the code output prediction task on the \href{https://livecodebench.github.io/}{LiveCodeBench} benchmark (see evaluation details in \ref{sec:codebench}). The data set contains multiple program implementations grouped into equivalence classes sharing the same functionality but differing in compositional structure. We create train-test splits with varying degrees of class leakage based on function names (heldout = 0, 0.5, 1.0), then fine-tune with LoRA and evaluate on test splits. We keep the total number of samples in the fine-tuning set constant across class-leakage values to control for sample-size effects. We treat test accuracy as a measure of compositional generalization performance, under the assumption that the pre-training data did not include training on code-output prediction tasks—an assumption we validate by reporting baseline performance for the non-fine-tuned model (No LoRA).

\textbf{Results:} In Figure~\ref{fig:LiveCodeBench_combined}(a), we observe that as we leak composition equivalences into training, it substantially increases overall accuracy (h=0.5, 1.0), whereas performance on held-out non-leaked functions (h=0.0) improves only minimally. This holds regardless of function-name anonymization, in which we replace function names with random strings to rule out the possibility that the model is simply learning to map function names to correct outputs. In Figure~\ref{fig:LiveCodeBench_combined}(b), we report per-function accuracy across different LoRA configurations. For most functions, leaking equivalences into training improves performance on the remaining held-out equivalences. For some functions (e.g., \texttt{distanceTraveled}), non-trivial performance is only achievable when equivalences are leaked. However, a small number of functions (e.g., \texttt{minLength}) show improvement even without equivalence leakage, which may be due to unaccounted equivalences—for instance, structurally similar functions operating on different data types with different function names. These results demonstrate that composition equivalences also affect compositional generalization in large pre-trained models on real-world tasks.

\section{Conclusion and Limitations}
We identify two key factors, composition equivalence and module coverage, that together explain substantial variability in compositional generalization across transformer variants. A key takeaway from our work is rethinking what ``unseen'' means in sequential compositional generalization: rather than a syntactic definition, we argue for a functional one---a test composition is truly out-of-distribution if it produces a novel end-to-end mapping relative to all training compositions. Our findings also reveal concrete failure modes in current evaluation protocols: benchmarks containing composition equivalences can significantly overestimate direct model generalization, and seemingly reasonable design choices (such as identity-token padding or combining within-$k$ and cross-$k$ evaluation) can be highly misleading. Benchmarks that include module coverage shifts and compositional equivalences can further exacerbate shortcut learning. We hope these findings motivate a more careful benchmark design to reliably assess compositional generalization.

\paragraph{Limitations:} Our findings are mainly established on sequential compositional tasks, and it remains an open question how composition equivalence and module coverage generalize to more complex compositional structures (e.g., hierarchical or graph-based compositions). In our analysis on LiveCodeBench, individual implementations sometimes differed substantially in their compositional structure due to the use of different libraries. Moving away from a synthetic setting means we can no longer control module coverage, length, order, or compositional structure. We therefore do not claim that compositional equivalence is the \textit{only} mechanism behind the observed performance gains; rather, the observed behaviors align strongly with our hypotheses about its effect on compositional generalization performance observed in real-world settings. Precisely isolating each factor requires either controlled synthetic settings---as we have done---or careful curation of real-world data sets with explicitly controlled train/test distribution shifts, which is an exciting future direction. 

\section*{Acknowledgements}
We thank Sankaran Vaidyanathan for the helpful comments and suggestions on an earlier version of the paper. 

\bibliography{references}

\appendix

\section{Appendix}

\subsection{Large-language model usage}
\label{app:llm_usage}
We have used large language models to minimally polish the writing in the main paper and to format the technical details in the Appendix.

\subsection{Related Work}
\label{ref:appendix:extended_rel_works}
In this section, we discuss additional related work.

\textbf{Other studies of compositional generalization in transformers:} A large body of work has studied compositional generalization in transformer-based models across a range of settings using both synthetic and real-world benchmarks. These works differ broadly in what constitutes a component—natural language instructions \citep{lake2018generalization, hupkes2020compositionality}, atomic skills \citep{NEURIPS2024_b99fd305}, data inputs for a fixed algorithmic task \citep{ahuja2025on}, and module functions in program executions \citep{ramesh2024compositional}—including in-context learning settings \citep{li2023transformers, garg2022can, xu2024large, an2023context}. In this work, a component generally refers to an atomic, reusable function that forms part of a sequential composition, similar to modules used in program execution. Our key focus is on task-based generalization \citep{ramesh2024compositional, abedsoltan2025task}, in which the labeling function varies with the module sequence, thereby requiring extrapolation to new tasks.

Several works \citep{petty2024impact, csordas2021devil, ontanon-etal-2022-making, zhang2024initialization} investigate the effects of different hyperparameters and architectural choices on transformer performance across various generalization tasks. Others \citep{li2023transformers, garg2022can, xu2024large, an2023context} have considered the compositional generalization capabilities of transformers when trained on samples using in-context learning. Several works \citep{li2023dissecting, wei2022chain, zhou2023leasttomost, lichain} investigate compositional generalization in the context of chain-of-thought prompts, reporting performance gains across various compositional tasks. \cite{yang2024exploring} evaluates the performance of compositional capabilities of large-language models after instruction-tuning by testing on unseen combinations of instructions. Another line of work investigates the internal circuitry learned to enable compositional generalization \citep{song2025out, wang2024grokking}, identifying specific parts of the transformer architecture that affect generalization. Our work contributes to this area of research by investigating the effect of module orderings and data-generating characteristics on the compositional generalization of transformer models. 

\textbf{Training bias and shortcut learning:}
Shortcut learning \citep{geirhos2020shortcut, duShortcutLearning} is the phenomenon where models rely on superficial features in the training data, which may have spurious correlations with the output, instead of the robust features that capture the true underlying data-generating process. We can view shortcut learning as one consequence of composition equivalence and module coverage violations. Understanding whether data exhibit compositional equivalence or violate module coverage provides a principled approach to mitigating spurious correlations and identifying cases where the model would exhibit robust out-of-distribution generalization.

\subsection{Module Function Definitions}
\subsubsection{Uniform Benchmark}
We define six random bijection functions $f: \mathcal{V}^m \rightarrow \mathcal{V}^m$ that randomly map each input data token $x_i \in \mathcal{V}:= \{a,b \ldots z\}$ to a random output data token $y_i \in \mathcal{V}:= \{a,b \ldots z\}$, based on a pre-defined lookup table. We ensure there is a one-to-one, unique mapping between input and output data tokens, and each input data token doesn't map to itself. We assume input data length $m=6$. The six tokens are sampled uniformly at random from $\mathcal{V}$ without replacement.

\subsubsection{Diverse Benchmark}
\label{appendix:task_token_definition}
\begin{enumerate}
    \item \texttt{shift(x)}: The shift function applies a predetermined \textit{bijective} transformation to each character in the input string according to a fixed character substitution table. \\ 
    \texttt{shift(h j f s d h) = i k g t e i}
    \item \texttt{sort(x):} The sorting function rearranges the characters of the input string in lexicographic (ascending alphabetical) order. \\
\texttt{sort(c g m a h b) = a b c g h m} 
    \item \texttt{mode(x):} The mode function returns the character that appears most frequently in the input string. In case of equal frequencies, the lexicographically smallest character is selected. \\
\texttt{mode(w j d n k k) = k}
    \item \texttt{concatenate(x, y):} The concatenation function performs string concatenation, appending the second string to the end of the first string.\\
 \texttt{join(s e w l r z, y s e o q n) = s e w l r z y s e o q n}
    \item \texttt{union(x, y):} The union function returns the ordered union of unique characters from both input strings, preserving the order of first occurrence within each string. \\ 
    \texttt{union(s e w l r z, y s e o q n) = s e w l r z y o q n}
    \item \texttt{filter(x):} The filter function extracts all vowel characters from the input string while preserving their original order. \\
\texttt{filter(d c f o j i) = o i}
    \item \texttt{identity(x)}: The identity function returns the input string unchanged.
    \texttt{identity(x) = x} 
\end{enumerate}

We sample two input data tokens of length six from the vocabulary $\mathcal{V} = \{a, b, c, \ldots z\}$, as some of the module functions are binary. The maximum possible length of output is $12$. We ignore the remaining data tokens if the length exceeds 12 to maintain a fixed output length. We pad with the space token in case the output is less than length 12, for example, in case of \texttt{mode}, \texttt {filter}, \texttt{union} functions.

\subsection{Transformer Architecture}
\label{transformer_architecture_details}

We use a modified \texttt{nanoGPT} implementation \cite{karpathy2023nanogpt} with three transformer layers, each containing a multi-head causal self-attention block (6 heads, embedding size 120), layer normalization, and an MLP with GELU activation. The model has a context window of 36 tokens for direct models and 114 tokens for step-by-step models. The vocabulary size is around 40 tokens. 
We disable dropout and biases in LayerNorm layers and apply weight tying between the token embedding and the output projection layer. Both absolute and relative (global) positional encodings were tested in separate experiments.  

Models are trained with an autoregressive objective, predicting the next token given the previous sequence. For a sequence $x_{1:T}$ of length $T$, the training loss is the cross-entropy:
\[
L(w) = - \sum_{t=1}^{T-1} \log p_w(y = x_{t+1} \mid x_{1:t}),
\]
where $p_w$ denotes the model distribution parameterized by weights $w$.

We train for 100 epochs with a batch size of 512. The optimizer is AdamW with $\beta_1 = 0.9$, $\beta_2 = 0.95$, and weight decay $0.1$. 
The learning rate follows a cosine annealing schedule with warmup 
(100 steps), starting from $3\times10^{-4}$ and decaying to 
$6\times10^{-6}$. Gradient clipping is applied with a maximum norm of 1.  Training is performed on a single GPU using PyTorch~2.0, with Flash Attention kernels (\texttt{scaled\_dot\_product\_attention}) enabled when available.

\subsubsection{Absolute and relative positional embeddings}
\label{positional_embeddings}

We experiment with two positional encoding embeddings: \textit{absolute} embeddings and \textit{relative global} embeddings.

\paragraph{Absolute embeddings.}  
In the absolute case, we follow the original transformer formulation 
\citep{vaswani2017attention}, where a learnable position embedding $P(t)$ is added to 
the token embedding $E(x_t)$ before being passed into the transformer layers:
\[
z_t = E(x_t) + P(t), \quad t = 1, \dots, T.
\]
This ensures that positional information is directly encoded in the input sequence representation, with the embedding table learned jointly with the model. In our implementation, the transformer instantiates both token and position embedding tables, which are added elementwise at each time step.

\paragraph{Relative global embeddings.}  
In the relative case, we replace the absolute embedding table with a learned relative representation incorporated into the attention mechanism. This approach is motivated by the relative position representations of \citet{shaw2018self} and extended in the 
Music Transformer \citep{anna2018music}. Formally, each attention head maintains a trainable relative embedding matrix $E_r \in \mathbb{R}^{C \times d_h}$, where $C$ is the maximum context length and $d_h$ the head dimension. Given query vectors $Q \in 
\mathbb{R}^{B \times H \times T \times d_h}$, we compute relative logits as:
\[
S_{\text{rel}} = \text{skew}(Q E_r^\top),
\]
where the skew operation aligns relative positions with their corresponding offsets. These logits are then added to the standard dot-product attention $QK^\top / 
\sqrt{d_h}$, yielding attention weights that encode both content and relative distance. This modification removes the absolute embedding table, instead parameterizing $E_r$ within each attention block.

\subsection{Training Details}
\label{app:experimental_details}

\textbf{Data generation.} We generate $100K$ training samples evenly distributed across the unique train permutations and $10k$ test samples evenly distributed across the unique test permutations. For example, for $k=6$, there are a total $720$ module permutations without identity module--so we sample roughly 138 samples per permutation with random data strings combined in train and test sets. All input data tokens are of fixed-size strings of length six and sampled uniformly without replacement from the $\mathcal{V} = \{a-z\}$. For binary functions in the diverse benchmark, we sample two input strings of length $6$, and the output can be of max-length $12$, from the same vocabulary space. As the uniform benchmark consists of unary bijection functions, both input and output data tokens are of length $6$.  

\textbf{Distinct input data strings:} As we randomly sample input strings of length six without replacement, there are a total of 11,576,560 unique strings. In our data sets, we found approximately 99.9\% unique strings across both the training and test sets of total size 110k (100k + 10k = 110k).
.

\textbf{Prompt format:} To generate the training sequences for the transformers, we serialize a vocabulary consisting of lowercase alphabet 
characters along with special tokens 
\texttt{<START>}, \texttt{<SEP>}, \texttt{<END>}, and a space token. \texttt{<START>} and \texttt{<END>} mark the sequence boundaries, while 
\texttt{<SEP>} separates function tokens, input strings, intermediate outputs (for step-by-step data), and the final output. Spaces are used for padding to ensure fixed input lengths across examples.  

We evaluate module orderings on two sets of functions. In the \textit{uniform} case, all functions are bijective map operations and require only a single input string. In the \textit{diverse} case, where functions are based on common string operations, functions may take up to two input strings and therefore each prompt must contain two input sequences. With this in mind, the following is the exact prompt structure used to train and evaluate our models:

For data points generated with the \textit{direct} setting, the prompt structure is:

\texttt{<START>} $f_1 f_2 \ldots f_k$ \texttt{<SEP>} $x_1 \ldots x_m$ \texttt{<SEP>} $x'_1 \ldots x'_m$ \texttt{<SEP>} $y_1 \ldots y_m$ \texttt{<END>}

where $f_1,\dots,f_k$ denote the module composition, $x_1\dots x_n$ the first input string, $[\texttt{<SEP>}\;x'_1\dots x'_n\,]$ the second input string present only in the diverse setting, and $y_1\dots y_m$ the final output.  

In the \textit{step-by-step} setting, the intermediate outputs of each function are 
also included between separators, yielding multiple \texttt{<SEP>} segments:

\texttt{<START>} $f_1 f_2 \ldots f_k$ \texttt{<SEP>}\; $x_1 \ldots x_m$ 
 \texttt{<SEP>} $x'_1 \ldots x'_m$ 
 \texttt{<SEP>} $y_{11} \ldots y_{1m}$ \texttt{<SEP>} $y_{21} \ldots y_{2m}$ \ldots \texttt{<SEP>}\; $y_{k1} \ldots y_{km}$
\texttt{<END>}

where $x^{(i)}$ denotes the intermediate output after applying the $i$th function in the composition. 

\textbf{Positional embeddings and length generalization:} Positional embeddings have been a major area of research when considering the length generalization of transformer architectures. Length generalization in transformers is the behavior where models trained on shorter sequences still perform well on longer ones unseen in the training data. In a sense, this requires that the transformer generalize seen modules at one position to perform the same task at another. This framing provided some of the motivation for including relative embeddings in transformers-the idea is that sequences of set behaviors should be learned relative to each other instead of at an absolute position. Indeed, several works \citep{neishi-yoshinaga-2019-relation, ruoss-etal-2023-randomized} have shown that using relative positional embeddings outperforms absolute ones. We refer readers to the survey by \cite{zhao-etal-2024-length} for a larger collection of community efforts. It is this same motivation that informs our decisions to consider relative positional embeddings. Though we do not test length generalization (since our input prompts are fixed length), we instead consider scenarios where modules do not appear in specific positions or appear less frequently.

\paragraph{Compute Resources:} All experiments were conducted on the
Unity SLURM cluster. NanoGPT models were trained on a single GPU with 100GB
RAM and 4 CPUs, with a wall-clock time of up to 1-2 hours per job. Gemma-3-1B and Llama-3-8B fine-tuning were performed on a single NVIDIA A40 GPU with 100 GB RAM and 8 CPUs, with a wall-clock time of up to 48 hours per job. Given the large number of train-test splits evaluated, experiments were parallelized across multiple SLURM jobs.

\subsubsection{Training, validation, and test performance during training}
\label{app:train_val_during_training}
\begin{figure}[h!]
    \centering
    \includegraphics[width=\linewidth]{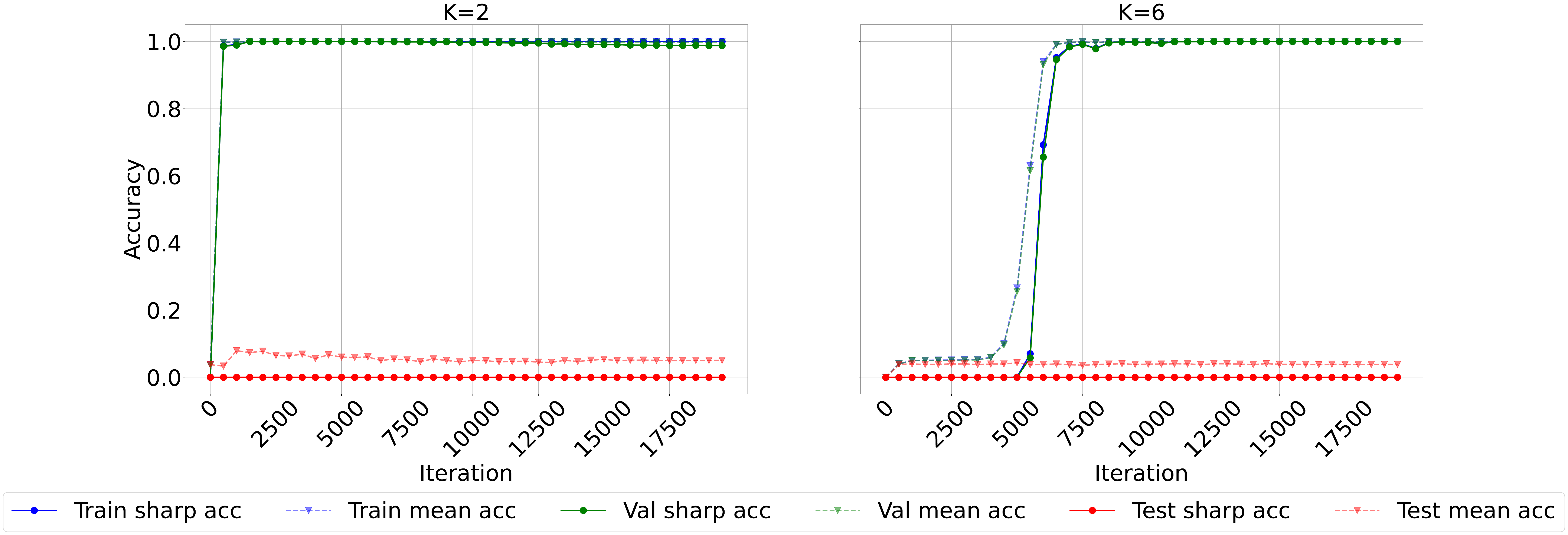}
    \caption{\textbf{Uniform benchmark (without identity), N\_heads = 6, N\_layers = 3, Direct (Relative)}:  We observe that the models learn to predict the outputs of training compositions for both K=2 and for K=6, while the test accuracy on unseen compositions remains 0\% throughout. We report two metrics: (1) \textbf{sharp accuracy}, where a prediction scores 1.0 only if the entire output sequence is correct and 0.0 otherwise; and (2) \textbf{mean token accuracy}, average accuracy of predictions in the output sequence, with both metrics averaged across compositions from each split. For reference, 100 epochs correspond to roughly 20k iterations (batch size 512, 100k training samples, 195 iterations per epoch).
 }
    \label{fig:equivalence_coverage_failures}
\end{figure}

\begin{figure}[h!]
    \centering
    \includegraphics[width=\linewidth]{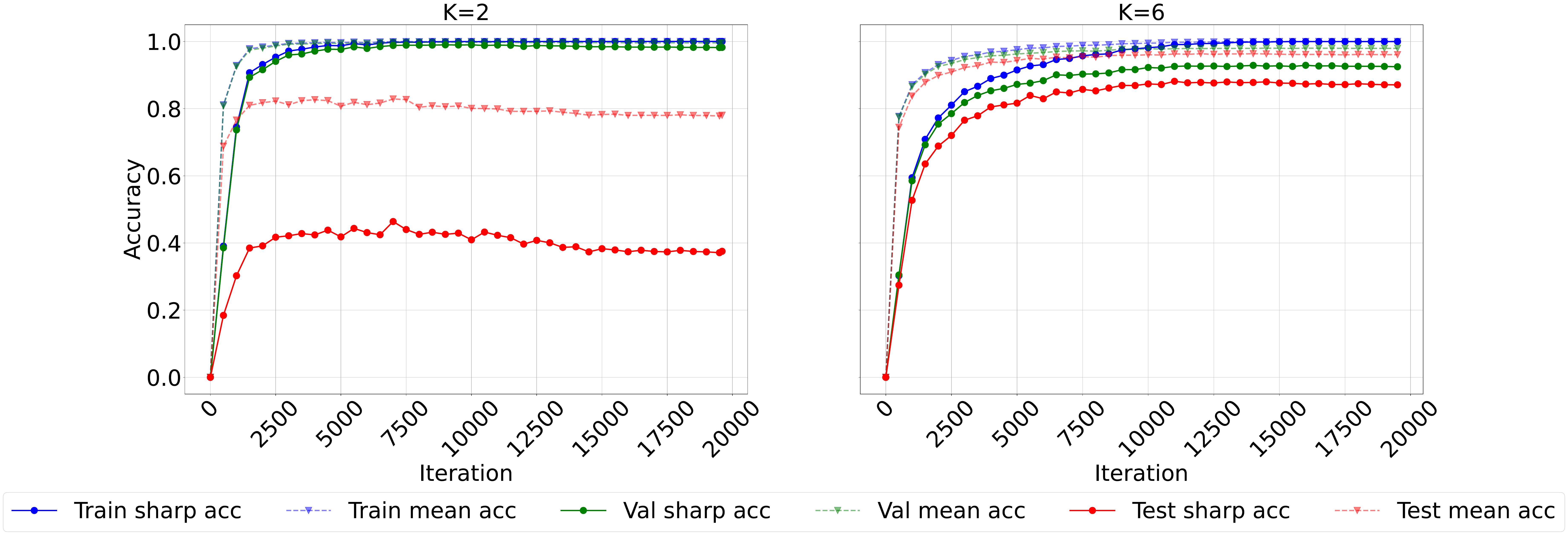}
    \caption{\textbf{Diverse benchmark, N\_heads = 6, N\_layers = 3, Direct (Relative)}: We observe non-trivial compositional generalization, as the models perform well on training and validation compositions, the test performance also increases. We observe better generalization for $K=6$ than for $K=2$ due to more training diversity and shared equivalences.}
    \label{fig:equivalence_coverage_failures}
\end{figure}

\subsubsection{Larger Transformer Architecture (N\_heads = 12, N\_layers=12)}
\label{app:larger_gpt2}
\begin{figure}[h!]
    \centering
    \includegraphics[width=\textwidth]{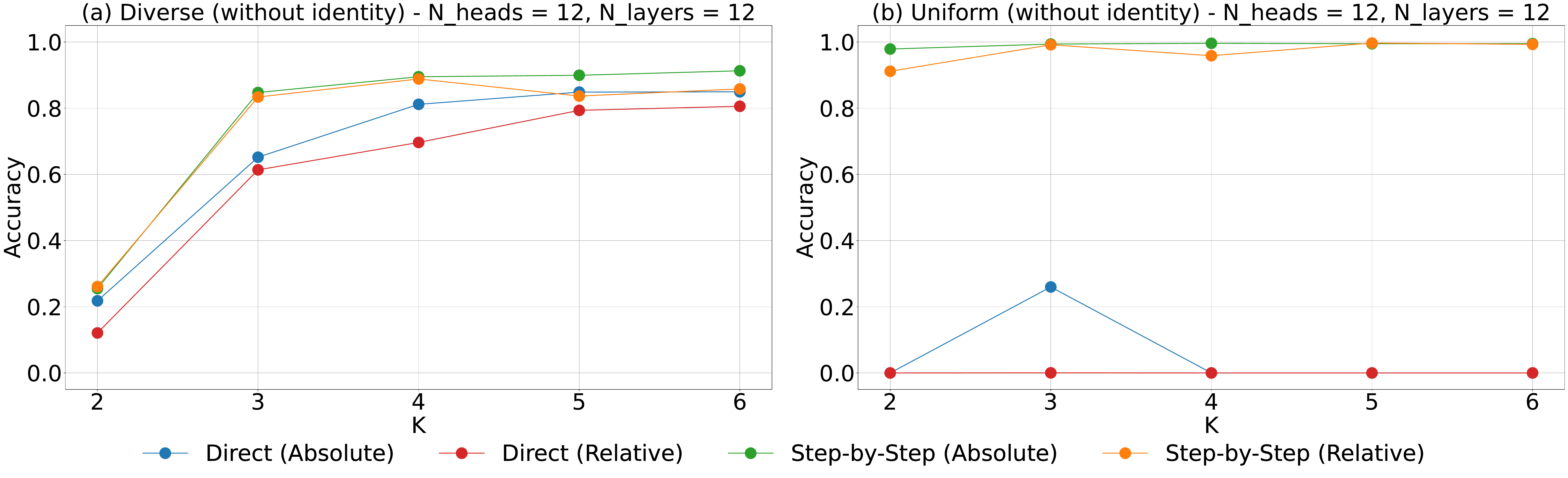}
    \caption{\color{black}\textbf{Generalization performance of bigger transformer architecture}:
We observe that the performance of the bigger architecture shows similar performance trends as seen for the smaller architecture (N\_heads = 6, N\_layers = 3) in Figures \ref{fig:identity_experiment}(b) and (d).}
\label{fig:nh12_nl12_architecture}
\end{figure}

In Figure \ref{fig:nh12_nl12_architecture}, we observe that the performance of the larger architecture shows similar trends seen in Figures \ref{fig:identity_experiment}(b) and (d) corresponding to the smaller architecture (N\_heads = 6, N\_layers = 3).

\subsection{Isolating the effect of composition equivalence on the performance of direct models}

\textbf{Sensitivity of results to $N$, the number of strings used to estimate composition equivalence:} Within each \texttt{num\_strings} setting, we consistently observe the same core trend reported in the paper: accuracy increases as shared equivalence (leakage) increases, regardless of $N$. Separately, $N$ trades estimation accuracy of the equivalence-class structure for compute cost. For splits with shared equivalence classes, larger $N$ tends to improve accuracy further, plausibly because test compositions are mapped more accurately to their correct equivalence class. Results are shown in Table~\ref{tab:num_strings_sensitivity}.

\begin{table}[t]
\centering
\caption{\textbf{Sensitivity of accuracy to the number of strings $N$ used to estimate composition equivalence.} Each column reports results for a given percentage of shared composition equivalences between train and test splits. Accuracy increases with shared equivalence at every value of $N$.}
\label{tab:num_strings_sensitivity}
\small
\setlength{\tabcolsep}{4pt}
\begin{tabular}{lcccccc}
\toprule
Num. strings & 0\% & 20\% & 50\% & 60\% & 70\% & 100\% \\
\midrule
10     & 1.8 & 4.2 & 6.4  & 12.7 & 12.3 & 83.4 \\
100    & 1.1 & 3.8 & 8.7  & 17.4 & 41.1 & 75.0 \\
1,000  & 0.4 & 6.2 & 10.0 & 22.0 & 44.9 & 78.2 \\
10,000 & 1.2 & 7.9 & 11.8 & 20.9 & 48.0 & 78.6 \\
\bottomrule
\end{tabular}
\end{table}

\begin{figure}[h!]
    \centering
    \includegraphics[width=\textwidth]{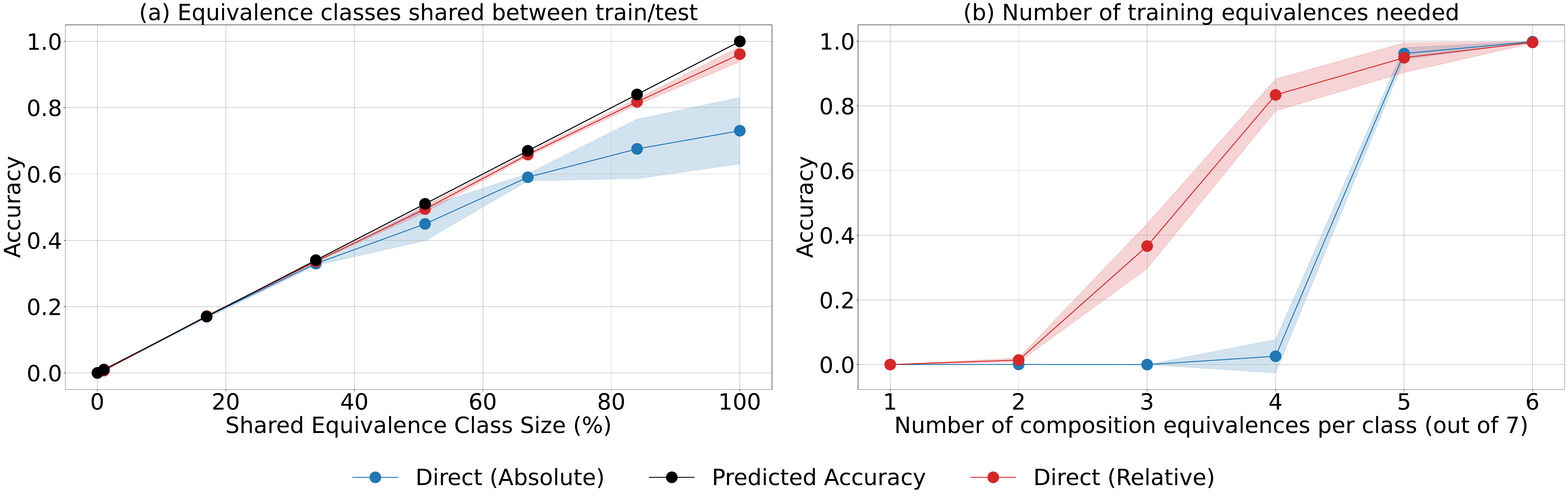}
    \caption{\textbf{Equivalence class necessity and training requirements:} (a) Generalization performance to unseen sequences is strongly correlated with the degree of shared composition equivalence classes between train and test. (b) Models need to see at least 4–5 composition equivalences (per class) in training to generalize to the corresponding test equivalences.}
    \label{fig:equivalence_class_leakage_experiment}
\end{figure}

\textbf{Uniform Benchmark:} Figure \ref{fig:equivalence_class_leakage_experiment}(a) shows that direct model performance, for both absolute and relative embeddings, is strongly predicted by the degree of shared equivalence classes. We systematically increase the proportion of shared identity-based equivalent classes in the train and test sets. In the uniform benchmark, there is a one-to-one mapping between sequences and identity-based equivalence classes, which simplifies controlling the number of shared equivalences. However, the mere existence of shared classes does not always guarantee correct generalization; for example, the absolute positional embedding model saturates in performance after 51\% shared classes in  Figure \ref{fig:equivalence_class_leakage_experiment}(a). Upon further analysis, we find that most failures correspond to test sequences that have the identity module in the first position (Figure \ref{fig:equivalence_coverage_failures}). 

\textbf{Details:} We focus on sequences with $k=6$ non-identity functions and one identity function, generating a total $7! = 5,040$ permutations grouped into a total of 620 equivalence classes. Each class consists of seven equivalent tasks corresponding to the possible positions of the identity module, while keeping the relative ordering of non-identity functions fixed.  
We maintain constant train-test split sizes in terms of equivalence classes (576/144) and vary the percentage of shared equivalence classes from 0–100\%. To maintain a fixed total number of samples, we exchange half of the tasks within shared equivalence classes between the training and test sets. A setting of 0\% means disjoint classes, while 100\% means all test tasks have equivalents in training. Accuracy is computed based on shared equivalence classes. 

\subsubsection{Equivalence class examples}

\begin{figure}
    \centering
    \includegraphics[width=0.9\linewidth]{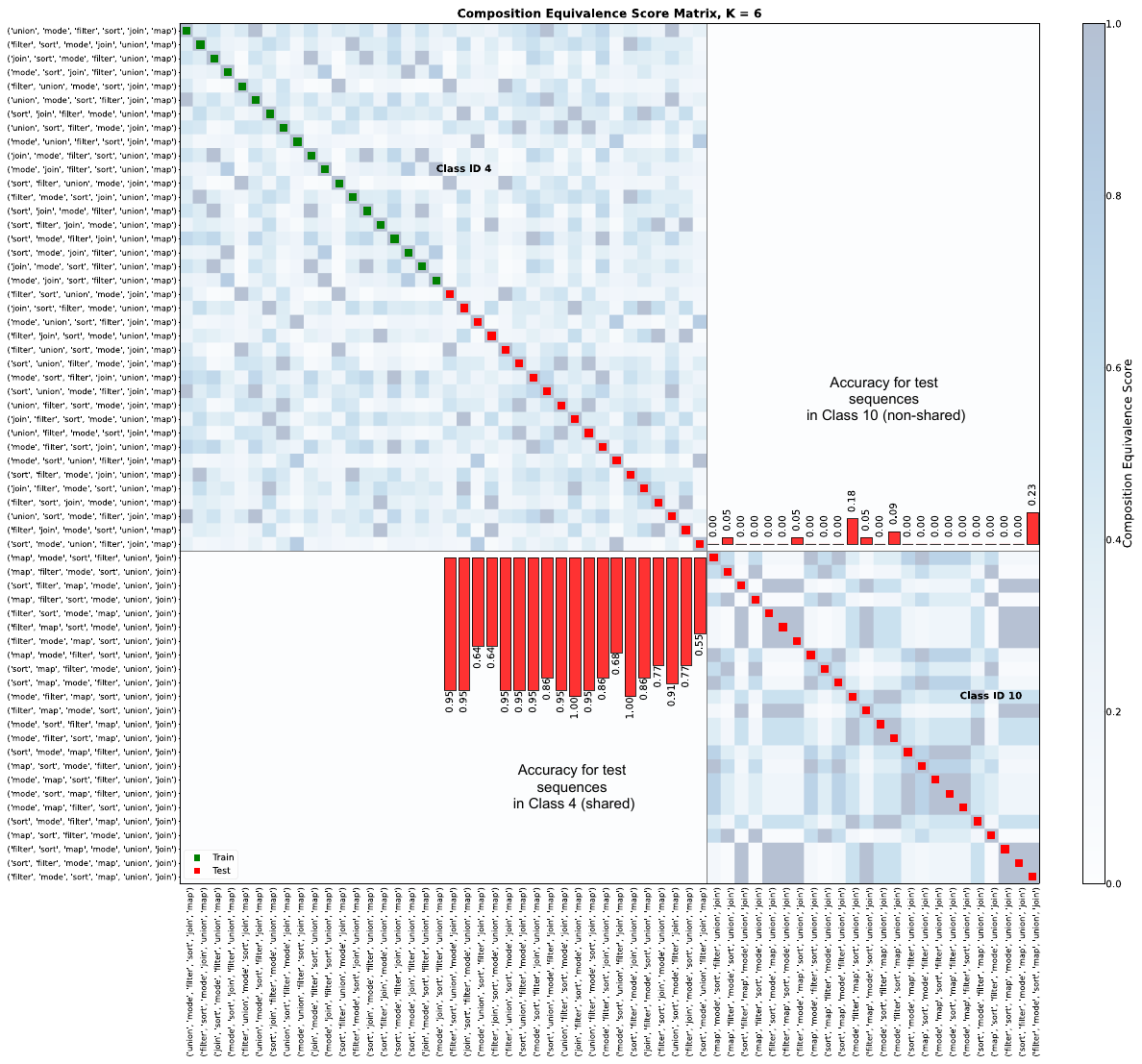}
    \caption{\color{black}\textbf{Visualization of learned equivalence classes for the diverse benchmark (K=6):} We show two equivalence classes (out of a total of 14 classes). First, we can note that the degree of equivalence varies significantly within an equivalence class. The test sequences are marked in red, and the training sequences are marked in green, indicating whether an equivalence class is shared. We observe that accuracy is significantly higher for compositions in the shared equivalence class and near zero for those in the non-shared equivalence class.}
    \label{fig:equivalence_class_visualization}
\end{figure}

\begin{figure}[h!]
    \centering
    \includegraphics[width=0.45\textwidth]{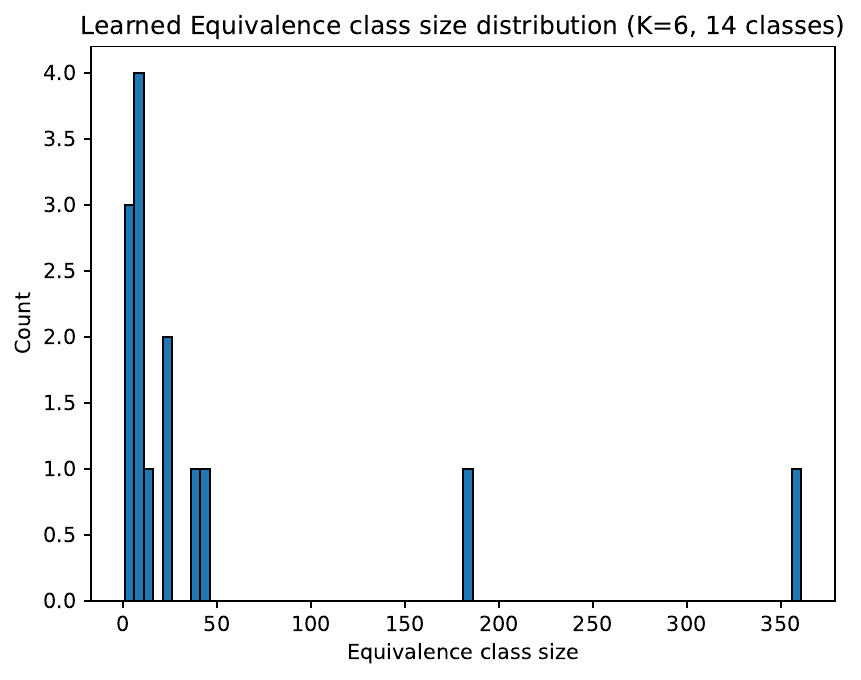}
    \caption{\color{black}\textbf{Size distribution of learned equivalence classes($K$ = 6)}}
\label{fig:size_distribution_eq_class}
\end{figure}

In Figure \ref{fig:equivalence_class_visualization}, we show two equivalence classes of size 38 (ID 9) and 24 (ID 2) for $K=6$. The first equivalence class usually ends with \texttt{(`union', 'map')} or \texttt{(`join', map')} sequences, and the second class ends with \texttt{(`union', 'join')}, causing end-to-end behavior difference between the two classes. We also observe that the equivalence scores vary significantly within an equivalence class, ranging from exact equivalence (1.0) to weak equivalence (0.01). Next, we observe that the model performs significantly better on test sequences belonging to equivalence classes with shared members in the train set and remains poor (0-0.2) on sequences belonging to non-shared equivalence classes. Figure~\ref{fig:size_distribution_eq_class} shows the size distribution of learned equivalence classes. 

\subsubsection{Systematically including test sequences without swapping with train sequences (K=6)}
\label{sec:without_swapping}

\begin{figure}[h!]
    \centering
    \includegraphics[width=\textwidth]{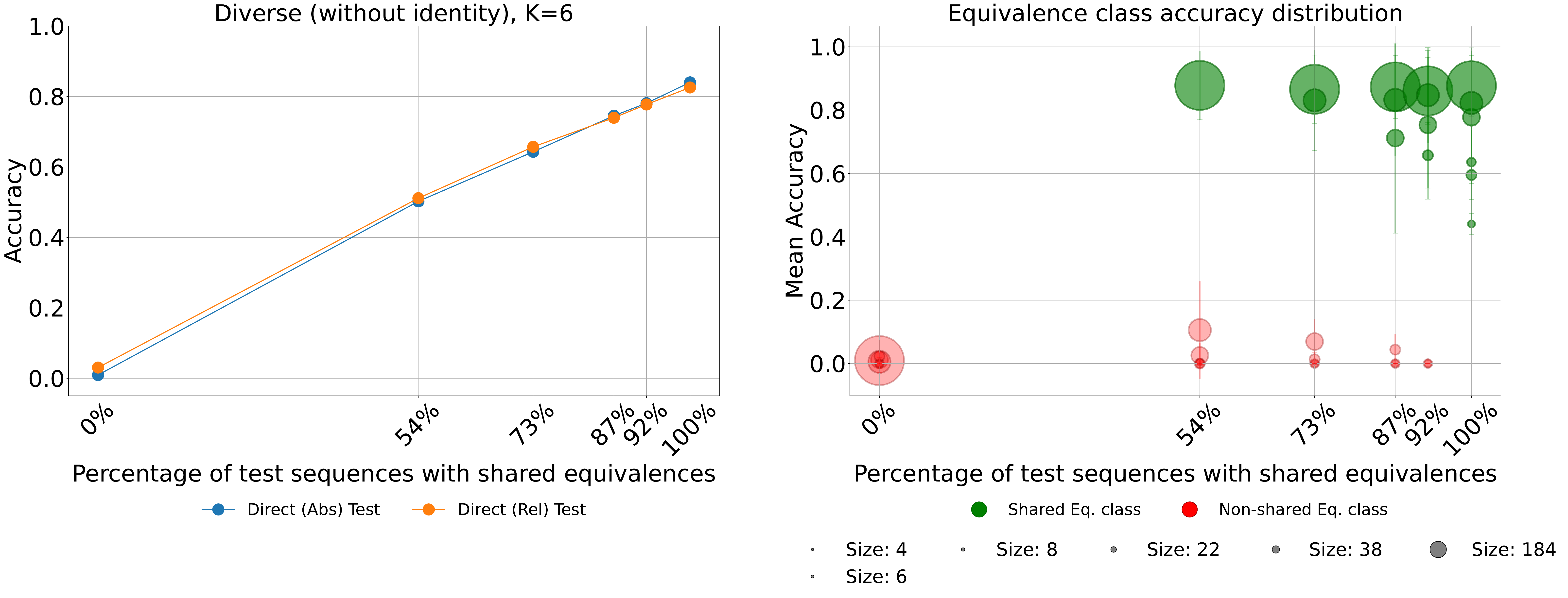}
    \caption{\color{black}\textbf{Leaking test equivalence classes without swapping ($K$ = 6):} (a) Generalization performance to unseen sequences is strongly correlated with the number of shared composition equivalence classes between train and test. (b) The mean accuracy of shared equivalence classes is significantly higher than that of non-shared equivalence classes.}
\label{fig:equivalence_class_leakage_experiment_without_swapping}
\end{figure}

In Figure \ref{fig:equivalence_class_leakage_experiment_without_swapping}(a), we show that the generalization performance increases with shared equivalences in the case we systematically leak half of the members of a test equivalence class in training without swapping. This reduces the test split size in proportion to the class size as the percentage of shared equivalences increases. Figure \ref{fig:equivalence_class_leakage_experiment_without_swapping}(b) shows that performance increases significantly for shared equivalence classes while performance of non-shared equivalences remains poor, varying between 0\%-20\%. 

\subsection{Composition equivalences encourage learning of shortcuts in step-by-step models}
\begin{figure}[H]
    \centering
    \begin{subfigure}[b]{0.48\textwidth}
        \centering
        \includegraphics[width=\textwidth]{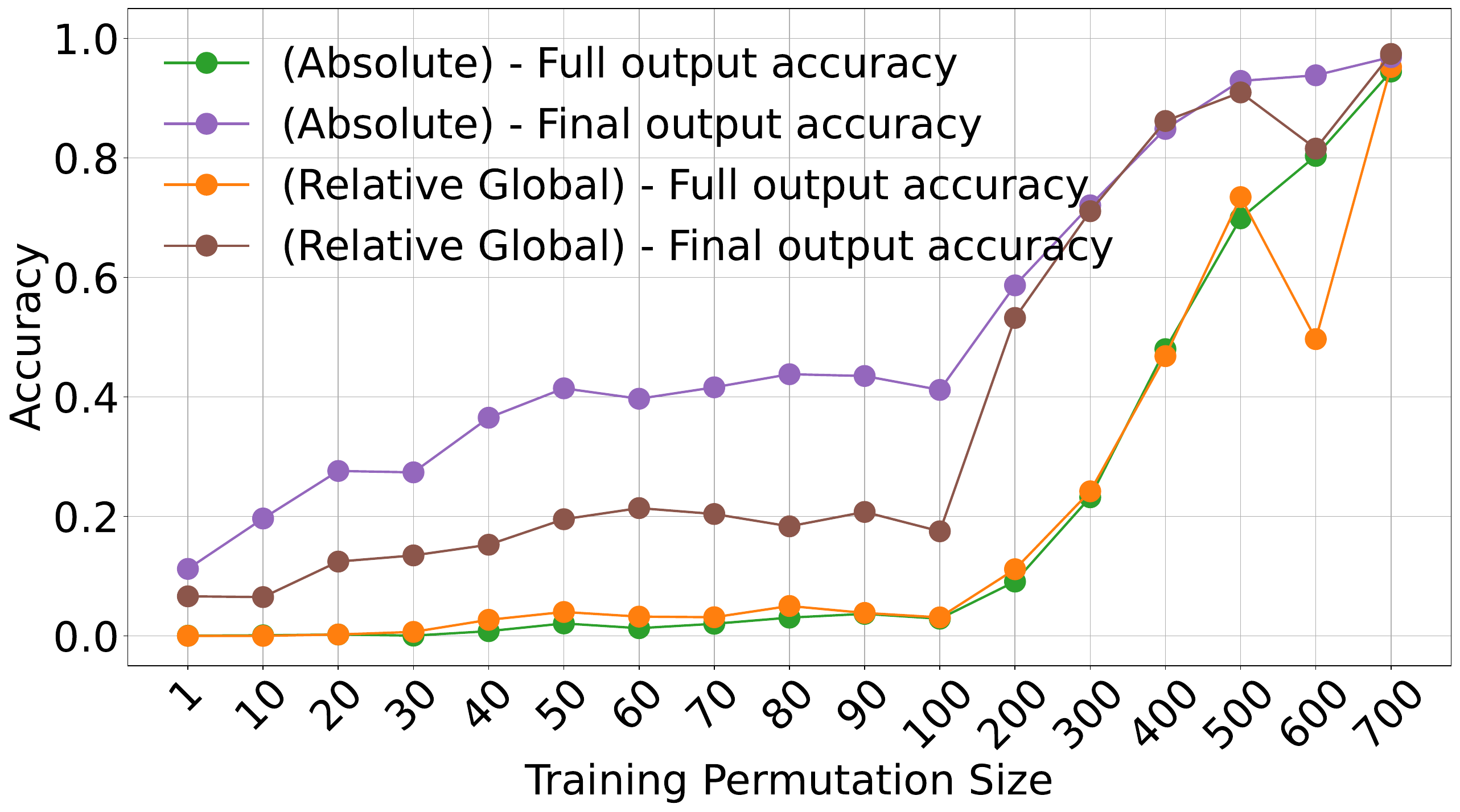}
        \caption{Within-$k$ evaluation ($k$= 6)}
        \label{fig:sbs_accuracy_uniform}
    \end{subfigure}   
    \hfill 
     \begin{subfigure}[b]{0.48\textwidth}
        \centering
        \includegraphics[width=\textwidth]{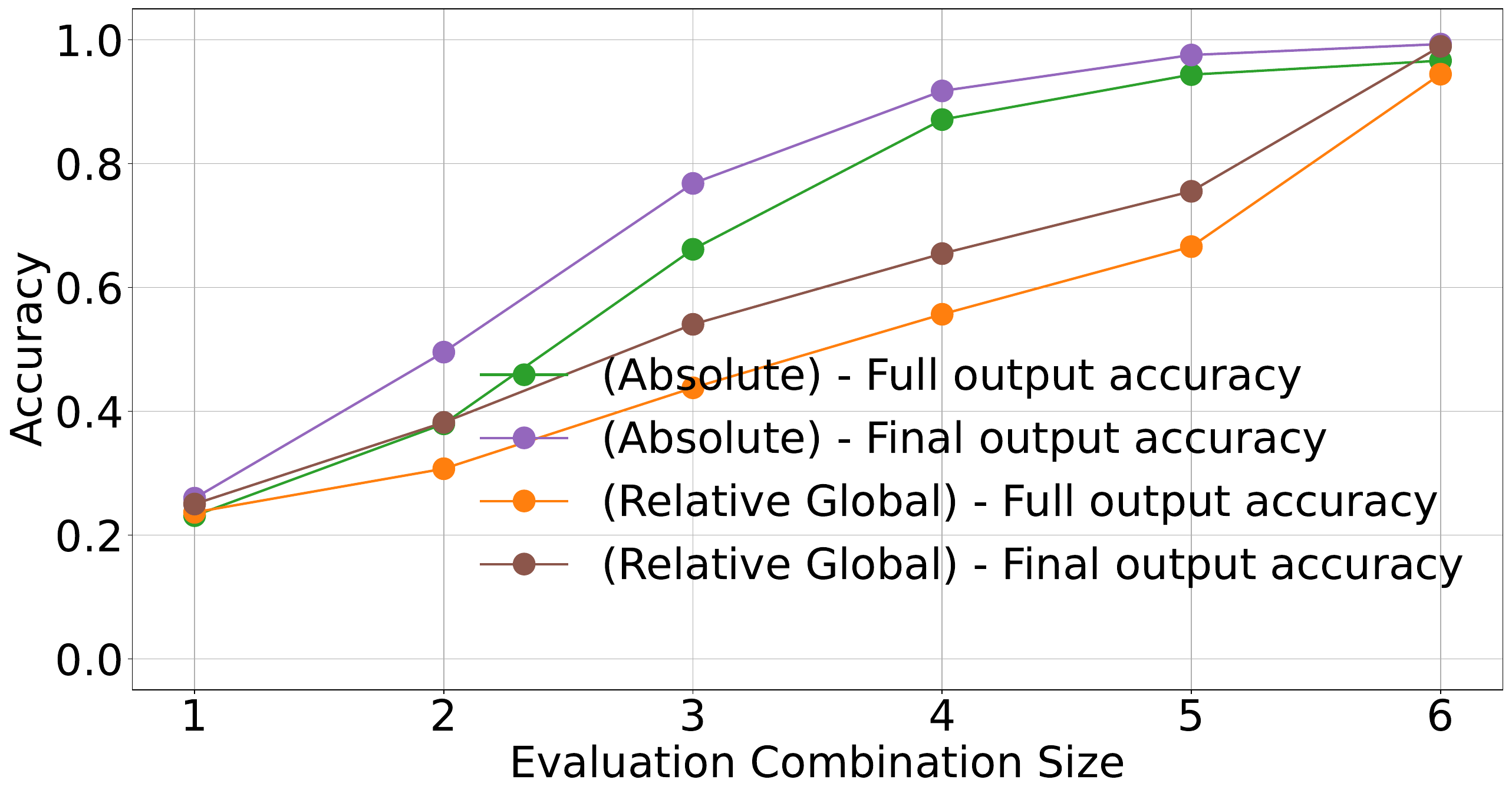}
        \caption{Cross-$k$ evaluation (Train-$k$ = 6)}
        \label{fig:sbs_accuracy_diverse}
    \end{subfigure}
    
    \caption{\textbf{Composition equivalence encourages shortcuts in step-by-step models:} We observe that composition equivalences in the diverse benchmark cause discrepancies between final output accuracy and full output (intermediate and final output) accuracy for both within-k evaluation and cross-k evaluation. Final accuracy $>$ full accuracy implies the model is learning shortcuts to reach the final answer. If we eliminate composition equivalences, the performance difference is zero between full and final outputs (see Figures \ref{fig:step_by_step_models_equivalence_uniform}(a), \ref{fig:step_by_step_models_equivalence_uniform}(b))}
    \label{fig:step_by_step_models_equivalence}
\end{figure}

\begin{figure}[h!]
    \centering
    \begin{subfigure}[b]{0.48\textwidth}
        \centering
        \includegraphics[width=\textwidth]{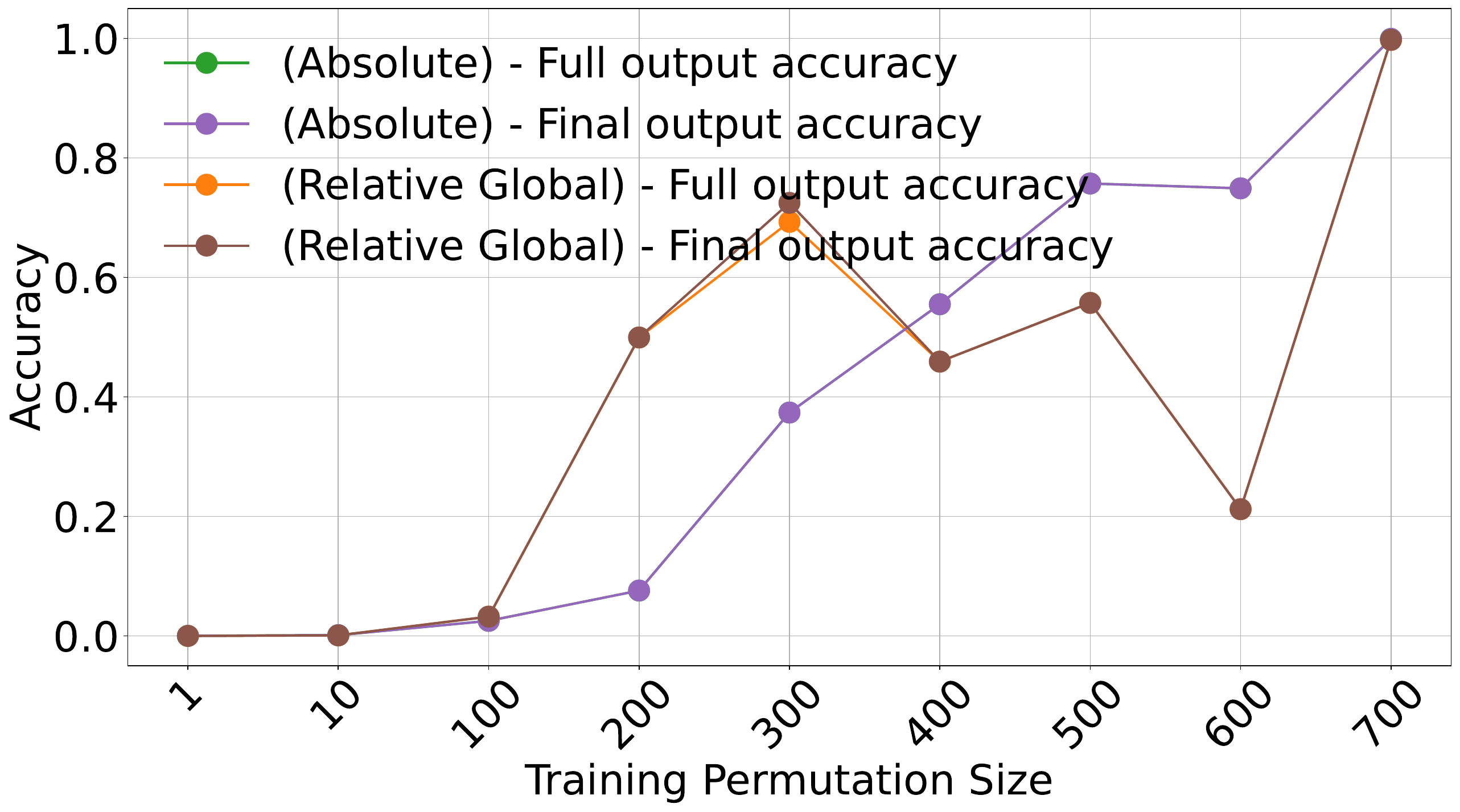}
        \caption{Within-$k$ evaluation ($k$= 6)}
        \label{fig:sbs_accuracy_uniform_within_k_shortcut}
    \end{subfigure}   
    \hfill 
     \begin{subfigure}[b]{0.48\textwidth}
        \centering
        \includegraphics[width=\textwidth]{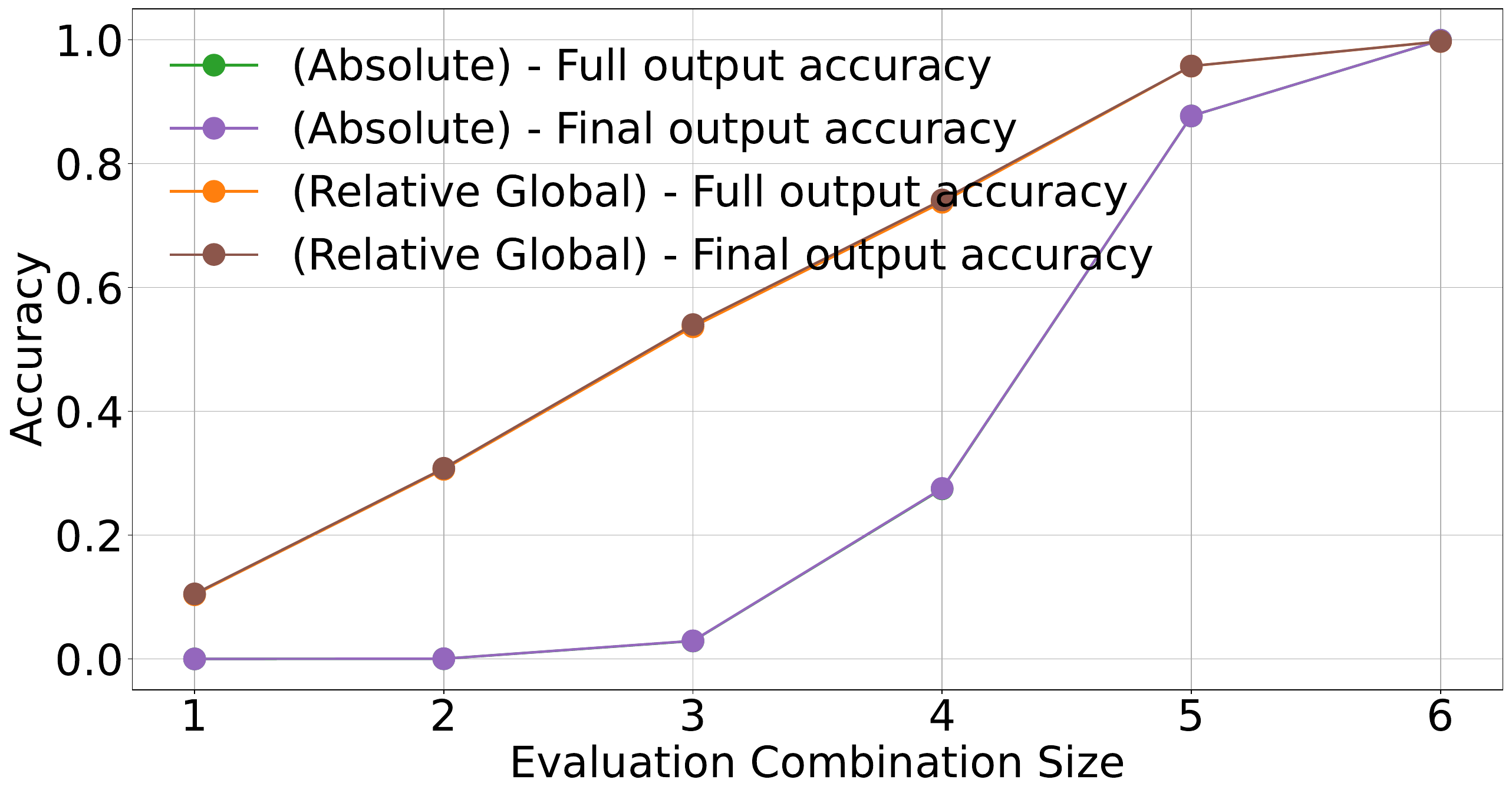}
        \caption{Cross-$k$ evaluation (Train-$k$ = 6)}
        \label{fig:sbs_accuracy_uniform_cross_k_shortcut}
    \end{subfigure}
    
    \caption{\textbf{Eliminating composition equivalences removes shortcut learning in step-by-step models :} We observe that the lack of compositional equivalence in the uniform benchmark removes shortcut-learning behavior in step-by-step models. This is shown by the exact same step-by-step accuracy, and the final accuracy is the same for both within-k evaluation ($k$ = 6) and cross-k evaluation (train-$k$ = 6). }
    \label{fig:step_by_step_models_equivalence_uniform}
\end{figure}

Training on the intermediate outputs breaks composition equivalences based on end-to-end labeling functions, improving the identifiability of individual labeling functions. Still, equivalences can also arise at the intermediate output level, especially when stepwise outputs are not distinct. We find that such equivalences encourage step-by-step models to rely on shortcuts rather than step-by-step reasoning, creating multiple paths to the correct answer. We observe that these models provide correct final outputs but incorrect intermediate ones in such cases. In the diverse benchmark consisting of a large number of equivalences, this results in higher final output accuracy but lower step-by-step accuracy, as models often produce incorrect intermediate outputs while still providing the correct final result (Figures~\ref{fig:step_by_step_models_equivalence}(a),(b)). In contrast, for the uniform benchmark, where no equivalences exist, step-by-step and final output accuracy exactly match as shown in Figures \ref{fig:step_by_step_models_equivalence_uniform}(a), \ref{fig:step_by_step_models_equivalence_uniform}(b). Thus, composition equivalence can \textit{negatively} impact compositional generalization in step-by-step models by promoting shortcuts over accurate step-by-step reasoning.

\subsection{The role of composition equivalence in compositional generalization of pre-trained models}
\label{sec:larger_models}
\subsubsection{Evaluation Details of Llama3-8B on LiveCodeBench benchmark:}
\label{sec:codebench}
\paragraph{Data set Curation} We perform LoRA fine-tuning \cite{hu2022lora} of pretrained \href{https://huggingface.co/meta-llama/Meta-Llama-3-8B}{Llama3-8B} to evaluate the performance of an LLM on a real-world task that may include compositional equivalences. Specifically, we train and evaluate on a dataset derived from LiveCodeBench's code execution evaluation task \cite{jain2025livecodebench}. LiveCodeBench extracts multiple solutions to problems hosted on various coding websites like LeetCode or CodeForces. Each individual solution can be considered compositionally equivalent to the others, since they have the same labeling function despite being implemented differently and sometimes using completely different libraries or methods. 

To evaluate the effect of the leakage of compositional equivalences, we select 9 functions which we use to evaluate the effect of equivalence leakage during fine-tuning (see Fig~\ref{fig:LiveCodeBench_combined}(b) for the exact function names). The functions were selected arbitrarily, with the only requirement being that they have the same number of code examples (6) in the original LiveCodeBench data set. From this, we generate a set of datasets with 100k training points and 10k testing points, varying the number of selected functions leaked into the training set, which we call the held-out fraction $h$. Leaked functions include four of the six code examples added to the pool of functions sampled for the training set. The remaining pool for both train and test is randomly sampled from a filtered set of functions from LiveCodeBench, which removes easy-to-predict functions (i.e., functions that output booleans) and functions that consistently throw errors when the inputs are randomized. For our experiments, we generate datasets for $h=0, 0.5, 1$, where we leak 0, 5, and 9 of the selected functions, respectively.

\paragraph{Fine-tuning Details} We fine-tune the model for 1 epoch using Low-Rank Adaptation with rank $r=16$ and scaling factor $\alpha=32$. A dropout rate of $0.05$ is applied to the LoRA updates, and no bias parameters are trained. LoRA is applied to the attention projection matrices $W_q, W_k, W_v,$ and $W_o$, while all pretrained model parameters are frozen.

\subsubsection{Gemma3-1B}
\label{sec:gemmaresults}
\begin{figure}[h!]
    \centering
    \includegraphics[width=\textwidth]{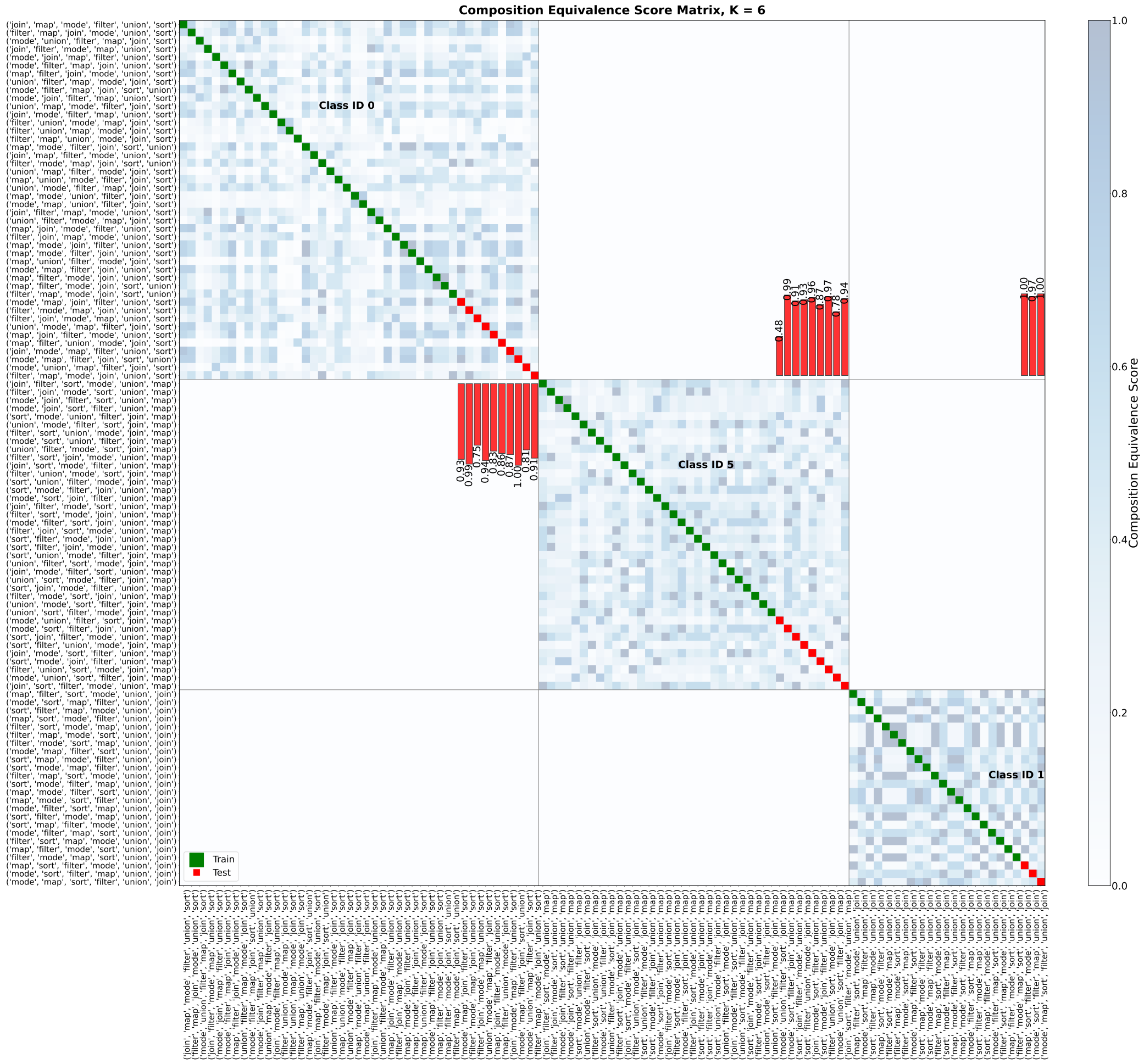}
    \caption{\color{black}\textbf{Gemma-1B visualization of per-class distribution for 100\% sharing:} We observe that most of the equivalence classes are shared and the model has overall high accuracy over test sequences in the shared equivalence classes.}
\label{gemma:comb_shared_examples}
\end{figure}

\begin{figure}[h!]
    \centering
    \includegraphics[width=\textwidth]{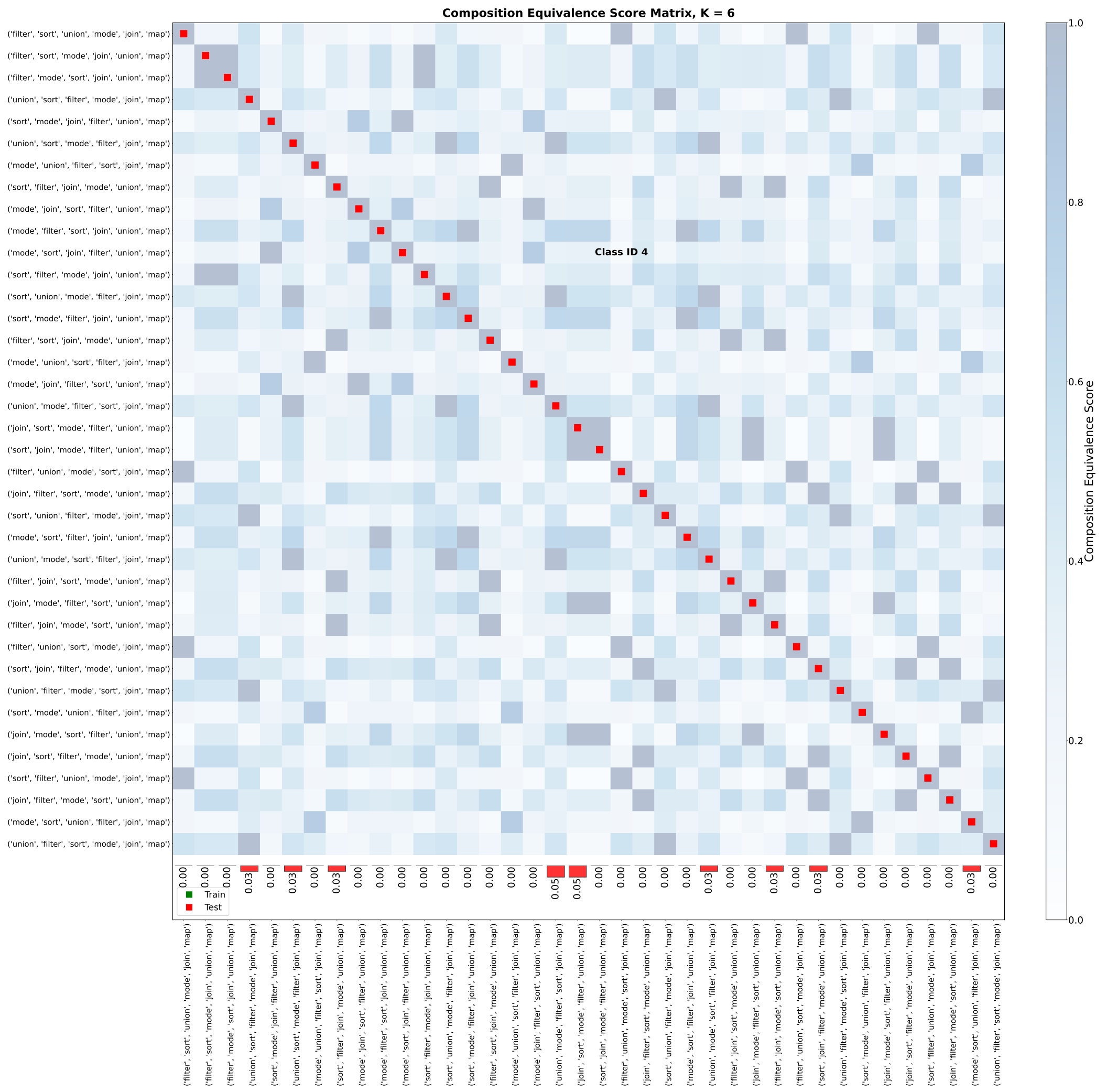}
    \caption{\color{black}\textbf{Gemma-1B visualization of test accuracy for the split with no shared equivalences:} We observe that none of the equivalence classes are shared, and the model has poor accuracy over test sequences in those classes.}
\label{gemma:disjoint_examples}
\end{figure}

We fine-tune and evaluate the compositional generalization performance of pre-trained \href{https://huggingface.co/google/gemma-3-1b-it}{Gemma3-1B} models on the following train/test splits. Fine-tuning is done for five epochs with a batch size of 1. Results can be seen in Figure~\ref{fig:equivalence_class_leakage_experiment_diverse}. Figure \ref{gemma:comb_shared_examples} shows the accuracy distribution over 100\% shared equivalences, and Figure \ref{gemma:disjoint_examples} shows the disjoint split, where no classes are shared, and the accuracy distribution of classes is pretty low. 

\subsection{Module coverage shift in cross-k settings}
\label{sec:cross_k_divergence}
\begin{figure}[h!]
    \centering
    \includegraphics[width=\linewidth]{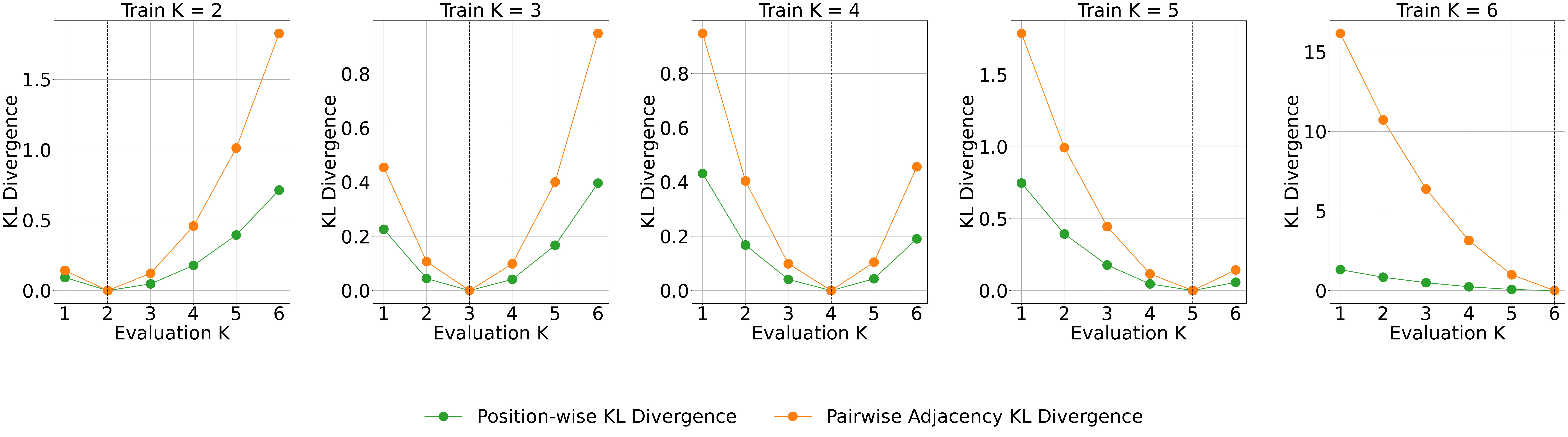}
    \caption{\textbf{Position-wise and pair-wise KL divergence in Cross-$k$ evaluation}}
    \label{fig:cross_k_divergence}
\end{figure}

In Figure \ref{fig:cross_k_divergence}, we show the KL divergence between train and test compositions as measured using shifts between position-wise and pair-wise module distributions. We can observe that the divergence increases as we move away from train $k$. The divergence values are correlated with the model performance. 

\subsection{Module coverage}
\label{app:mod_coverage_shift}
\subsubsection{Generation of various degrees of position-wise and pair-wise distribution shift}
\label{sec:pos_degree_shift}

\begin{definition}[\textbf{Position-wise module distribution}]
For module position $j \in \{1, 2, 3, \dots, k\}$ and module token $t_i \in \mathcal{T}$, the position-wise coverage distribution is: $\hat{P}(T_j = t_i) = \frac{1}{N} \sum_{s=1}^N \mathbb{I}(T_j^{(s)} = t_i).$
\end{definition}

\begin{definition}[\textbf{Pairwise module adjacency distribution}]
The empirical pairwise adjacency distribution is: $\hat{P}(T = t_i, T' = t_{i'}) = \frac{1}{N(k-1)} \sum_{s=1}^N \sum_{j=1}^{k-1} \mathbb{I}(T_j^{(s)} = t_i \text{ and } T_{j+1}^{(s)} = t_{i'}).$
\end{definition}

Given $n$ modules $\mathcal{T} = \{t_1, t_2, \ldots, t_n\}$ and composition length $k$, we generate train and test distributions with controlled divergence:

\textbf{Position-wise shift generation:}
\begin{enumerate}
    \item Begin with uniform distributions:
\begin{align}
P_{\text{train}}^{(0)}(T_j = t_i) = P_{\text{test}}^{(0)}(T_j = t_i) &= \frac{1}{n} \\
\forall j \in \{1, \ldots, k\}, \, \forall t_i \in \mathcal{T} \nonumber
\end{align}

\item Introduce bias strength $\alpha \in [0, 1]$ where $\alpha = 0$ yields no shift and $\alpha = 1$ yields maximum shift.

\item For each position $j$, define offset $\text{offset}(j) = j \bmod n$ and create complementary groups:
\begin{align}
\mathcal{G}_j^{\text{train}} &= \{t_{(i + \text{offset}(j)) \bmod n} : i = 0, \ldots, \lceil n/2 \rceil - 1\} \\
\mathcal{G}_j^{\text{test}} &= \mathcal{T} \setminus \mathcal{G}_j^{\text{train}}
\end{align}

\item Define un-normalized weights:
\begin{equation}
w_{\text{train}}(T_j = t_i; \alpha) = \begin{cases}
1 + 10\alpha & \text{if } t_i \in \mathcal{G}_j^{\text{train}} \\
1 - 0.95\alpha & \text{if } t_i \in \mathcal{G}_j^{\text{test}}
\end{cases}
\end{equation}
with $w_{\text{test}}$ defined symmetrically, then normalize to obtain $P_{\text{train}}(T_j = t_i; \alpha)$ and $P_{\text{test}}(T_j = t_i; \alpha)$.

\item Compute position-wise divergence: $D_{\text{pos}}(\alpha) = \frac{1}{k}\sum_{j=1}^k D_{\text{KL}}(P_{\text{test}}(T_j; \alpha) \| P_{\text{train}}(T_j; \alpha))$.
\end{enumerate}

\subsubsection{Effect of pairwise distribution shifts}
\begin{figure}[H]
    \centering
    \includegraphics[width=\linewidth]{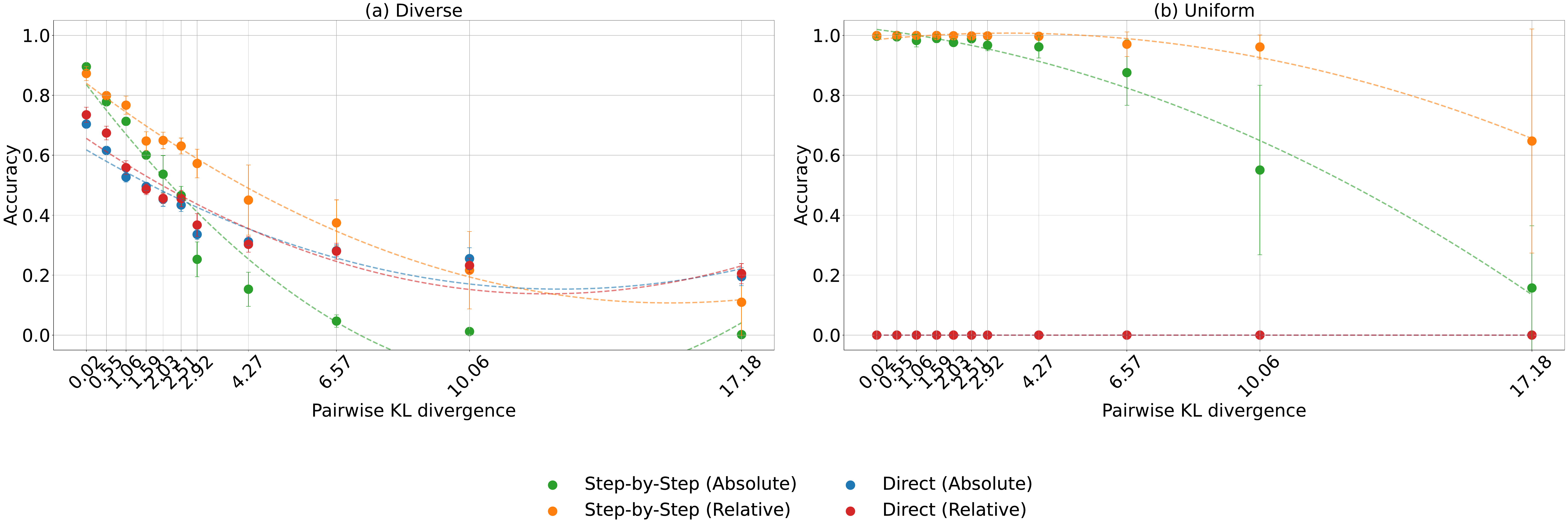}
    \caption{\textbf{Effect of pairwise-wise distribution shift:} (a) As the pairwise-wise KL divergence increases, the compositional generalization performance of both direct and step-by-step models decreases in the diverse benchmark. (b) For the uniform benchmark, the effect of divergence is primarily observed at the extreme right end for support violations.}
    
    \label{fig:pairwise_kl}
\end{figure}

In Figure \ref{fig:pairwise_kl}, we demonstrate the effect of pairwise distribution shift. Biased sampling of pairwise distributions is done in a similar way to the position-wise distribution. 

\textbf{Pairwise shift generation:}
Similarly, for adjacent module pairs $(T_j, T_{j+1})$, we define complementary pairwise groups $\mathcal{G}_{j}^{\text{train, pair}}$ and $\mathcal{G}_{j}^{\text{test, pair}}$ containing module pairs, apply analogous biased weighting with strength $\alpha$, and compute pairwise divergence: $D_{\text{pair}}(\alpha) = D_{\text{KL}}(P_{\text{test}}(T, T'; \alpha) \| P_{\text{train}}(T, T'; \alpha))$.

For both cases, we sample $N_{\text{train}}$ and $N_{\text{test}}$ compositions ensuring valid permutations and disjoint supports: $\text{supp}(\mathcal{P}_{\text{train}}^{\mathcal{T}}) \cap \text{supp}(\mathcal{P}_{\text{test}}^{\mathcal{T}}) = \emptyset$. Varying $\alpha \in \{0, 0.1, \ldots, 1.0\}$ systematically controls coverage shift magnitude.

\begin{figure}[h!]
    \centering
    \includegraphics[width=0.7\linewidth]{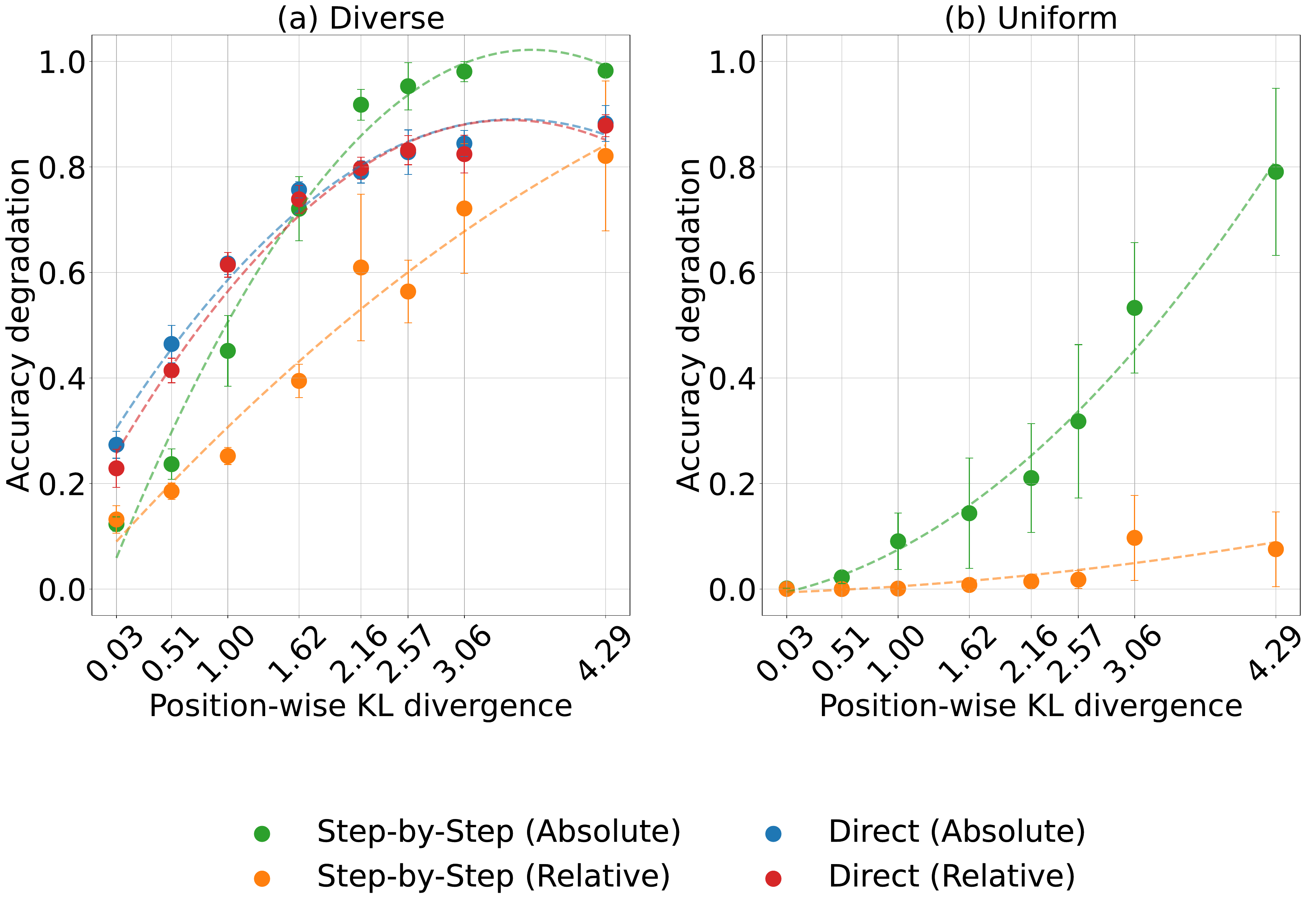}
    \caption{\textbf{Combined effect of position-wise shift and reduced diversity:} Shows (1-$accuracy_{KL}$) to demonstrate accuracy degradation as in Figure~\ref{fig:position_wise_kl}, revealing combined effects of coverage shift and reduced compositional diversity due to biased sampling. Here, $accuracy_{KL}$ is the performance with biased sampling.}
    \label{fig:pos_wise_diversity_controlled_a}
\end{figure}

\begin{figure}[h!]
    \centering
    \includegraphics[width=0.7\linewidth]{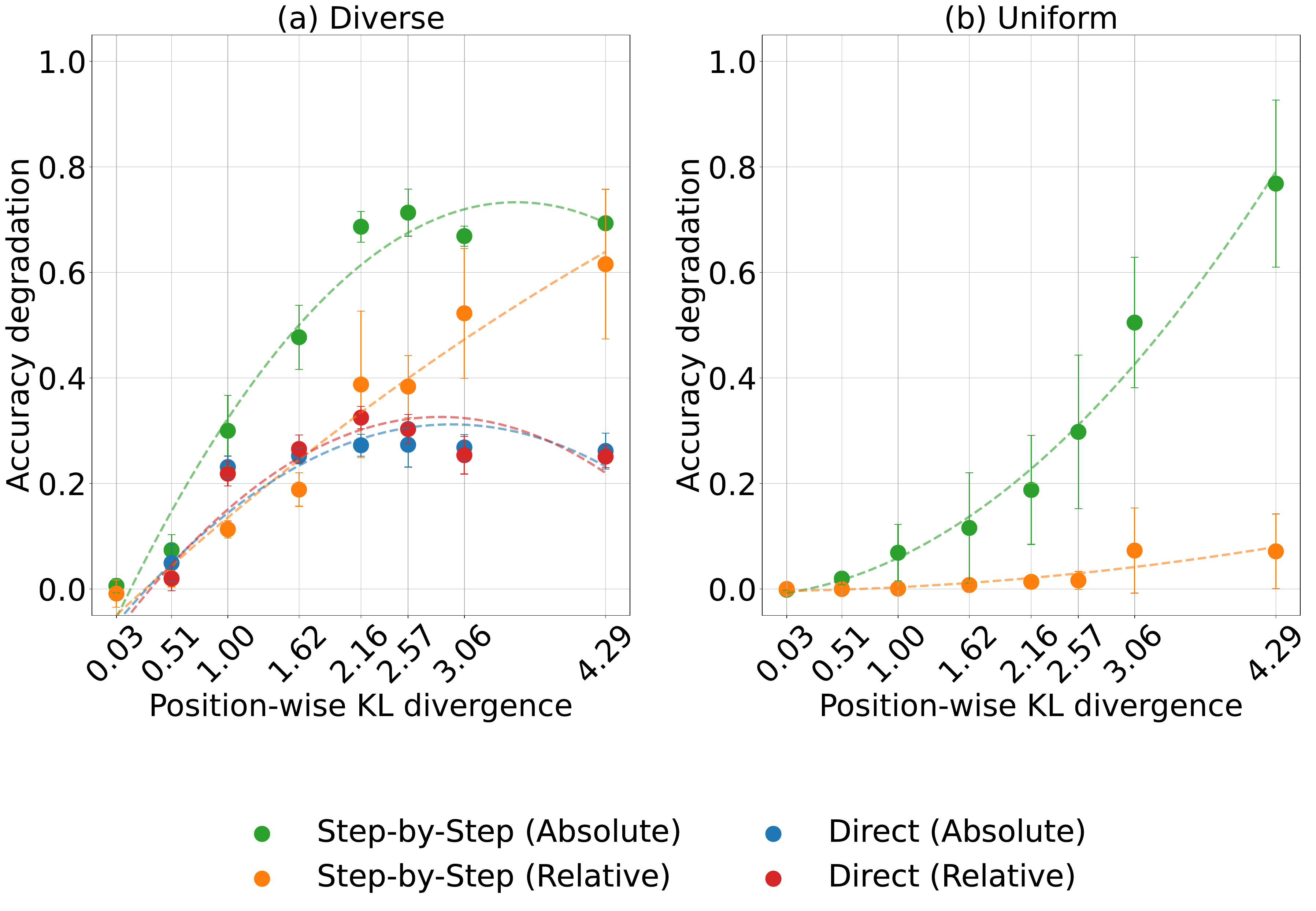}
    \caption{\textbf{Isolating position-wise shift by controlling diversity:} 
Shows $accuracy_{zero\_shift}-accuracy_{KL}$, where $accuracy_{zero\_shift}$ 
and $accuracy_{KL}$ are accuracies under random and biased sampling with $n$ 
unique compositions at each $D_{pos}$, respectively. For the diverse benchmark, 
position-wise shift alone has a reduced but still significant effect compared 
to Figure~\ref{fig:pos_wise_diversity_controlled_a}, indicating that both 
coverage shift and reduced diversity contribute to degradation. For the uniform 
benchmark, the negligible difference confirms that degradation is driven 
primarily by coverage shift alone.}
    \label{fig:pos_wise_diversity_controlled_b}
\end{figure}

\paragraph{Interplay between coverage shift and compositional diversity:}
If we allow sampling duplicate compositions, increasing bias strength causes certain compositions to be sampled more frequently. For target size $N=360$, increasing bias strength from $0$ to $1$ reduces unique compositions from $360$ to $50$ while increasing their sampling frequency. Consequently, Figure~\ref{fig:pos_wise_diversity_controlled_a} shows the combined effects of both module coverage shift and reduced compositional diversity. This interplay reflects realistic settings where limited diversity naturally induces coverage distribution shifts between train and test sets. To isolate these effects, we ran control experiments generating the same number of unique compositions observed at each $D_{pos}$ value, but with zero KL divergence via random sampling. Subtracting this zero-shift baseline isolates the coverage shift effect while controlling for diversity (Figure~\ref{fig:pos_wise_diversity_controlled_b}). We observe that the effect of coverage shift remains significant, though reduced compared to Figure~\ref{fig:pos_wise_diversity_controlled_a}, as expected.

\subsection{Sample efficiency ablations}
\label{app:sample_effeciency}
\begin{figure}[H]
    \centering
    \includegraphics[width=\linewidth]{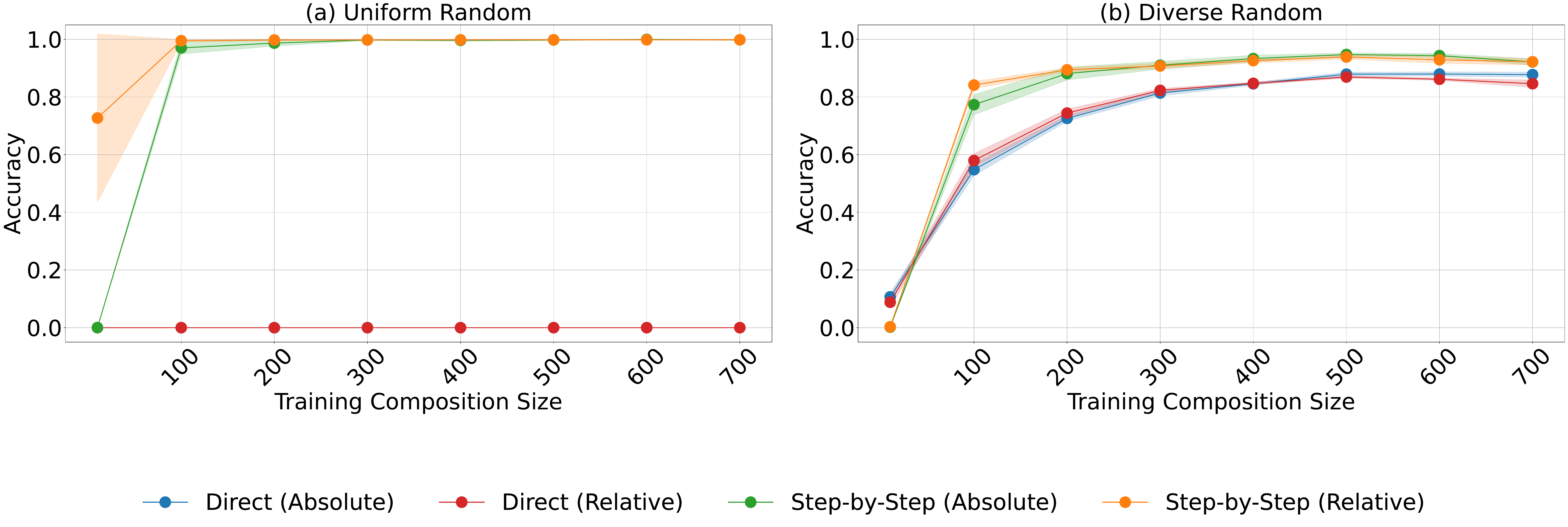}
    \caption{\textbf{Effect of number of unique training sequential compositions (randomly sampled, K=6):}: We observe that (a) step-by-step models with relative global embedding need fewer training compositions to generalize faster than absolute embedding models, and (b) For the diverse benchmark, as compared to direct models, step-by-step models need fewer unique training compositions due to extra information available using intermediate outputs.}
    \label{fig:equivalence_coverage_failures}
\end{figure}

\begin{figure}[H]
    \centering
    \includegraphics[width=\linewidth]{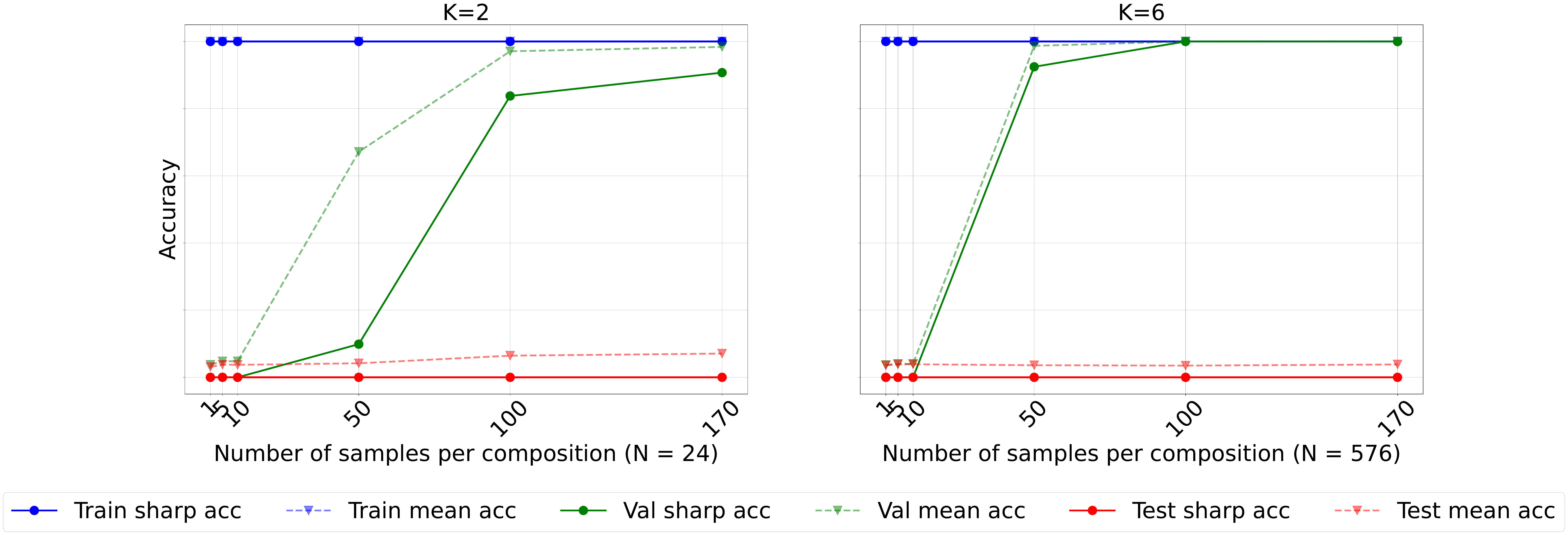}
    \caption{\textbf{Effect of number of samples per composition (uniform benchmark, direct model):}. Number of training compositions $N=24$ for $K=2$, and $N= 576$ for $K=6$. We vary the number of input-output samples per training composition in this experiment.  We observe that with a small number of training samples, the train/val gap is large, and the model struggles to generalize to training compositions with novel inputs. The train-val gap reduces as the number of samples increases. Test accuracies remain 0\%, as expected.}
    \label{fig:equivalence_coverage_failures}
\end{figure}

\begin{figure}[H]
    \centering
    \includegraphics[width=\linewidth]{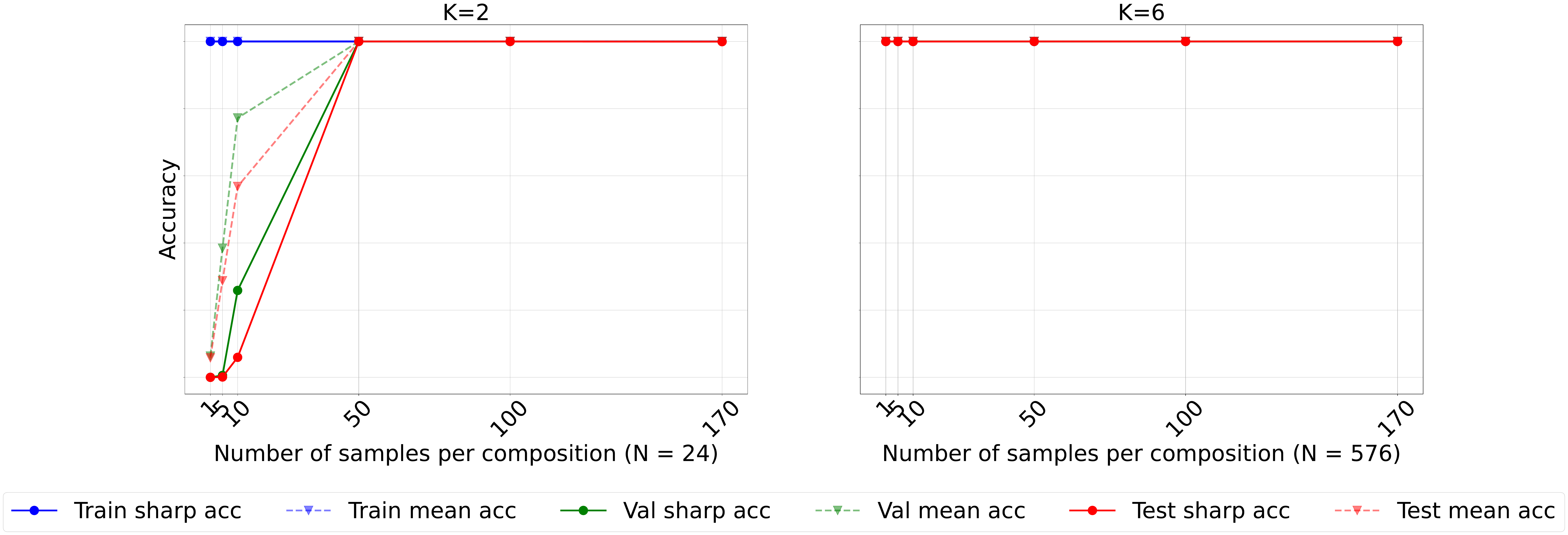}
    \caption{\textbf{Effect of number of samples per composition (uniform benchmark, step-by-step model):} We observe that step-by-step models are more sample efficient than direct models due to access to intermediate outputs providing extra information.}
    \label{fig:equivalence_coverage_failures}
\end{figure}

\begin{figure}[H]
    \centering
    \includegraphics[width=\linewidth]{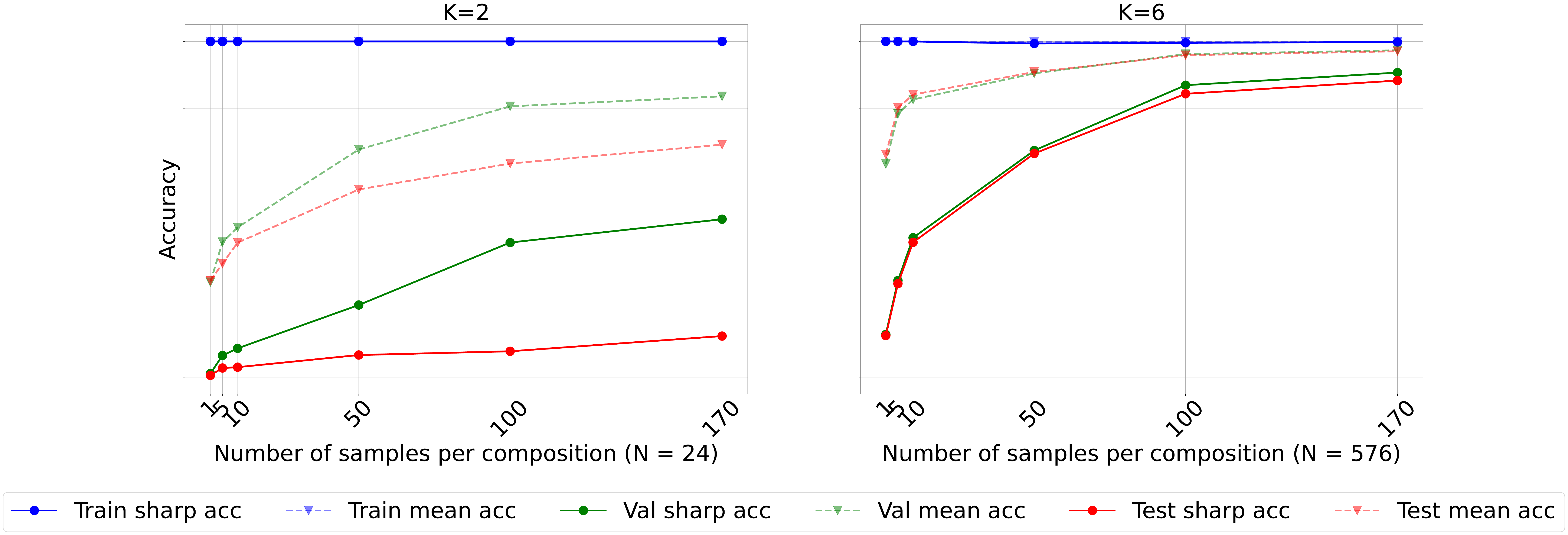}
    \caption{\textbf{Effect of number of samples per composition (diverse benchmark, direct model):} For the diverse benchmark, direct models are less sample efficient than the uniform benchmark to generalize on the validation set, especially for K= $2$. We observe significant compositional generalization behavior, as seen in Figure 2(b) for large samples, especially for K=$6$. }
    \label{fig:equivalence_coverage_failures}
\end{figure}

\begin{figure}[H]
    \centering
    \includegraphics[width=\linewidth]{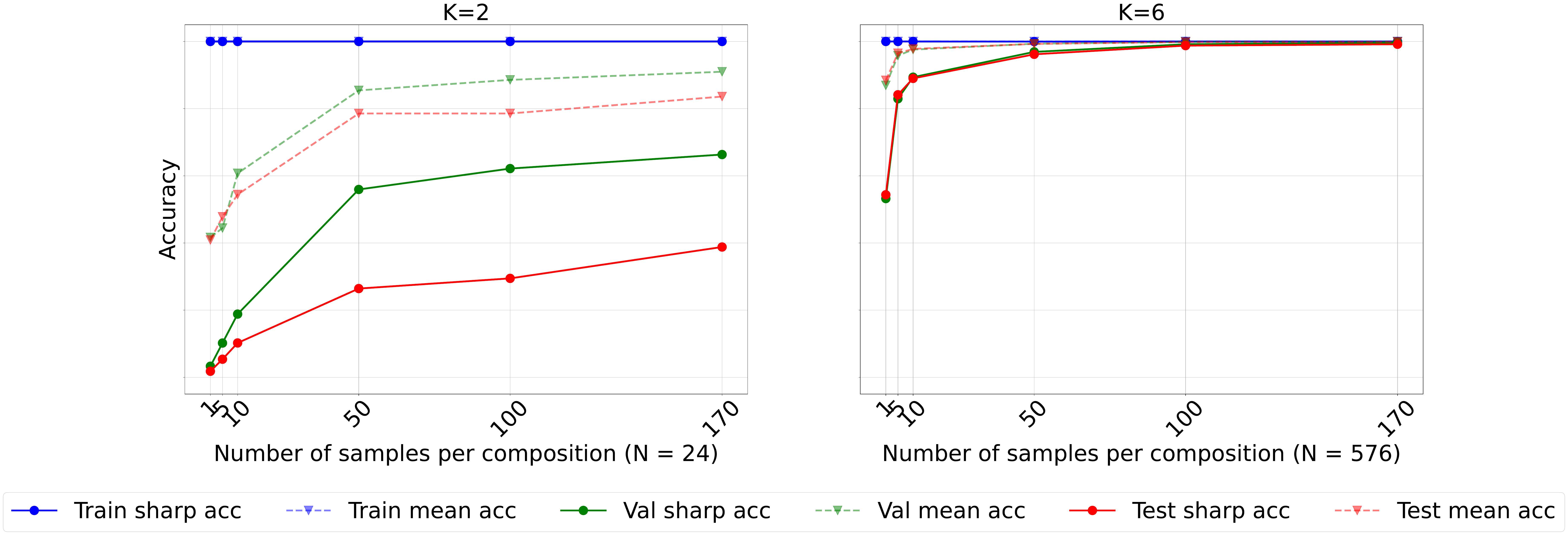}
    \caption{\textbf{Effect of number of samples per composition (diverse benchmark, step-by-step model):} Step-by-step models are more sample efficient than direct models, and a similar difference between uniform and diverse benchmarks for K=$2$ is observed as for direct models. }
    \label{fig:equivalence_coverage_failures}
\end{figure}
\newpage

\newpage

\end{document}